\documentclass[11pt,a4paper]{article}              

\usepackage[margin=25mm]{geometry}
\usepackage{graphics}
\usepackage{colortbl}
\usepackage{caption}
\usepackage{subfig}
\usepackage{graphicx}
\usepackage{fancyhdr}
\usepackage{amssymb,amsfonts,amsmath,amsthm}
\usepackage{tabularx}
\usepackage{natbib}
\usepackage{adjustbox}
\usepackage{multirow}
\usepackage{rotating}

\usepackage{makecell}

\usepackage{abbrevs}

\usepackage{authblk}
	
\graphicspath{{./},{./Figures/}}

\newcommand{\rev}[1]{ #1 }

\begin{document}
\title{Rare event estimation using stochastic spectral embedding} 

\author[1]{P.-R. Wagner} 
\author[1]{S. Marelli}  
\author[2]{I. Papaioannou} 
\author[2]{D. Straub} 
\author[1]{B. Sudret} 

\affil[1]{Chair of Risk, Safety and Uncertainty Quantification,
	ETH Zurich, Stefano-Franscini-Platz 5, 8093 Zurich, Switzerland}

\affil[2]{Engineering Risk Analysis Group, Technical University Munich, Arcisstraße 21, 80290 Munich, Germany}

\date{22.12.2021}
\maketitle

\abstract{
Estimating the probability of rare failure events is an essential step in the reliability assessment of engineering systems. Computing this failure probability for complex non-linear systems is challenging, and has recently spurred the development of active-learning reliability methods. These methods approximate the limit-state function (LSF) using surrogate models trained with a sequentially enriched set of model evaluations. A recently proposed method called stochastic spectral embedding (SSE) aims to improve the local approximation accuracy of global, spectral surrogate modelling techniques by sequentially embedding local residual expansions in subdomains of the input space. In this work we apply SSE to the LSF, giving rise to a stochastic spectral embedding-based reliability (SSER) method. The resulting partition of the input space decomposes the failure probability into a set of easy-to-compute \rev{conditional} failure probabilities. We propose a set of modifications that tailor the algorithm to efficiently solve rare event estimation problems. These modifications include specialized refinement domain selection, partitioning and enrichment strategies. We showcase the algorithm performance on four benchmark problems of various dimensionality and complexity in the LSF.\\[1em] 

}

\maketitle

\section{Introduction}
\label{sec:Introduction}

Ensuring the reliability of structures and systems is a core task in many engineering disciplines. This includes the reliability of machines \citep{Bertsche2008}, medical devices \citep{Fries2017}, electronic systems \citep{LaCombe1999} and civil engineering systems \citep{Haldar2006}. With the increasing availability of computer simulations, and continuous developments in numerical methods, model-based reliability analysis has become the state-of-the-art in reliability assessment. In this context, the task of reliability analysis lies in estimating the probability that an engineering system fails or performs undesirably. The computation of this so-called \emph{failure probability} is the objective of quantitative model-based reliability methods.

In reliability problems, the system under consideration is typically parameterized by a vector of random input parameters $\vX\colon\Omega\to\cD_{\vX}\subseteq\mathbb{R}^M$ following a probability distribution $\vX \sim f_{\vX}$. To identify the safe/failure state of a system as a function of the input parameters, a so-called \emph{limit-state function} $g\colon\mathcal{D}_{\vX}\to\mathbb{R}$ is introduced. It encodes the system performance \wrt the limit states that apply to the system under consideration (\eg ultimate stresses or maximum displacements at critical locations). Importantly, the function $g$ depends on one or several, often computationally demanding, engineering models. By convention, if $g(\vx) \le 0$, then $\vx$ corresponds to a failed configuration, otherwise it corresponds to a safe configuration. The interface between the resulting \emph{safe} and \emph{failure} domains is known as the \textit{limit-state surface}.

One can then define the probability of failure $P_f$ as:
\begin{equation}
	\label{eq:failureProb}
	P_f \eqdef \mathbb{P}\left[g(\vX)\le 0\right] = \int_{\cD_{\vX}} \indfun{\cD_f}(\vx)f_{\vX}(\vx)\,\di{\vx},
\end{equation}
where the indicator function $\indfun{\cD_f}(\vx)$ takes the value $1$ in the failure domain $\cD_f$ and $0$ everywhere else:
\begin{equation}
	\label{eq:failureInd}
	\indfun{\cD_f}(\vx) =
	\begin{cases}
		1, \quad \text{if} \quad g(\vx) \le 0,\\
		0, \quad \text{if} \quad g(\vx) > 0.
	\end{cases}
\end{equation}

Analytical computation of the failure probability is rarely possible in practice and direct numerical integration (\eg via quadrature) is often hindered by the inherently small scale of the failure probability and the potentially high input dimensionality $M$. For these reasons, initial efforts in the reliability literature focused on developing methods that approximate the limit-state surface \citep{Basler1960, Hasofer1974,Rackwitz78, Zhang2, Hohenbichler1987, Breitung1994,Tvedt90, Cai1994267}. These approximation methods remain competitive for a certain class of reliability problems even today, but there exist well known examples where the shape of the limit-state surface leads to gross errors in the estimated failure probability (\eg high-dimensional, non-linear problems \citet{Valdebenito2010a}). In those cases the direct estimation of the failure probability by means of probabilistic simulation methods is the method of choice. The most widely used methods for this are \emph{Monte Carlo simulation} (MCS, \citet{Fishman2011, Rubinstein2016}), approximation methods in conjunction with \emph{importance sampling} (IS, \citet{Hohenbichler1988, Melchers1999}), its \emph{adaptive variants} (sequential IS, \citet{Papaioannou2016} and cross entropy-based IS, \citet{Kurtz2013, Papaioannou2019a}) and \emph{subset simulation} (SuS, \citet{Au2001, Zuev2012}). Other noteworthy methods that can be used to directly estimate the failure probability are \emph{radial-based importance sampling} (RBIS, \citet{Harbitz1986}), \emph{directional simulation} (DS, \citet{Bjerager1988}) and \emph{line sampling} (LS, \citet{Koutsourelakis2004, Papaioannou2021}).

Computationally expensive models remain challenging to analyse, even with the most efficient simulation methods. In this setting, cheap to evaluate surrogate models can lead to significant computational savings, especially with so-called \emph{active learning reliability} methods. These methods approximate the limit-state function $g$ using a sequence of increasingly accurate surrogate models. The first method in this family was \emph{efficient global reliability analysis} (EGRA, \citet{Bichon2008}). Arguably the best known method today is \emph{adaptive Kriging} MCS (AK-MCS, \citet{Echard2011}) and its variants \citep{Echard2013, Huang2016}. A recent review of the field is given in \citet{Moustapha2021}.

In this paper we propose a novel active learning reliability method that utilizes the recently proposed \emph{stochastic spectral embedding} method (SSE, \citet{Marelli2020}) and more specifically the active learning sequential partitioning approach developed for Bayesian inverse problems in \citet{Wagner2020JCP}. The proposed approach, termed \emph{stochastic spectral embedding-based reliability} (SSER), benefits from a handful of modifications to the original SSE, namely new \emph{refinement domain selection}, \emph{partitioning} and \emph{sample enrichment} schemes, \rev{which we present in Section~\ref{sec:Methodology}. SSER constructs a set of spectral representations of $g$ with higher accuracy in proximity of the limit-state surface. The proposed adaptive partitioning strategy results in a set of subdomains in the vicinity of the limit-state surface. Provided that $g$ is sufficiently well-behaved in these subdomains, our approach is capable of efficiently addressing problems of varying complexity (\ie non-linear computational models, high dimensional problems, small failure probabilities). We showcase this property by comparing SSER to other state-of-the-art methods on four applications in Section~\ref{sec:Applications}.}

\section{Stochastic spectral embedding-based reliability}
\label{sec:Methodology}

\subsection{Stochastic spectral embedding}
\begin{figure}
	\centering
	\subfloat[Quantile space]{
		\label{fig:cs1:SSEBehaviour:1}
		\begin{minipage}{0.5\linewidth}
			\centering
			\includegraphics[width=0.7\linewidth, trim = 0 10 0 12 clip]{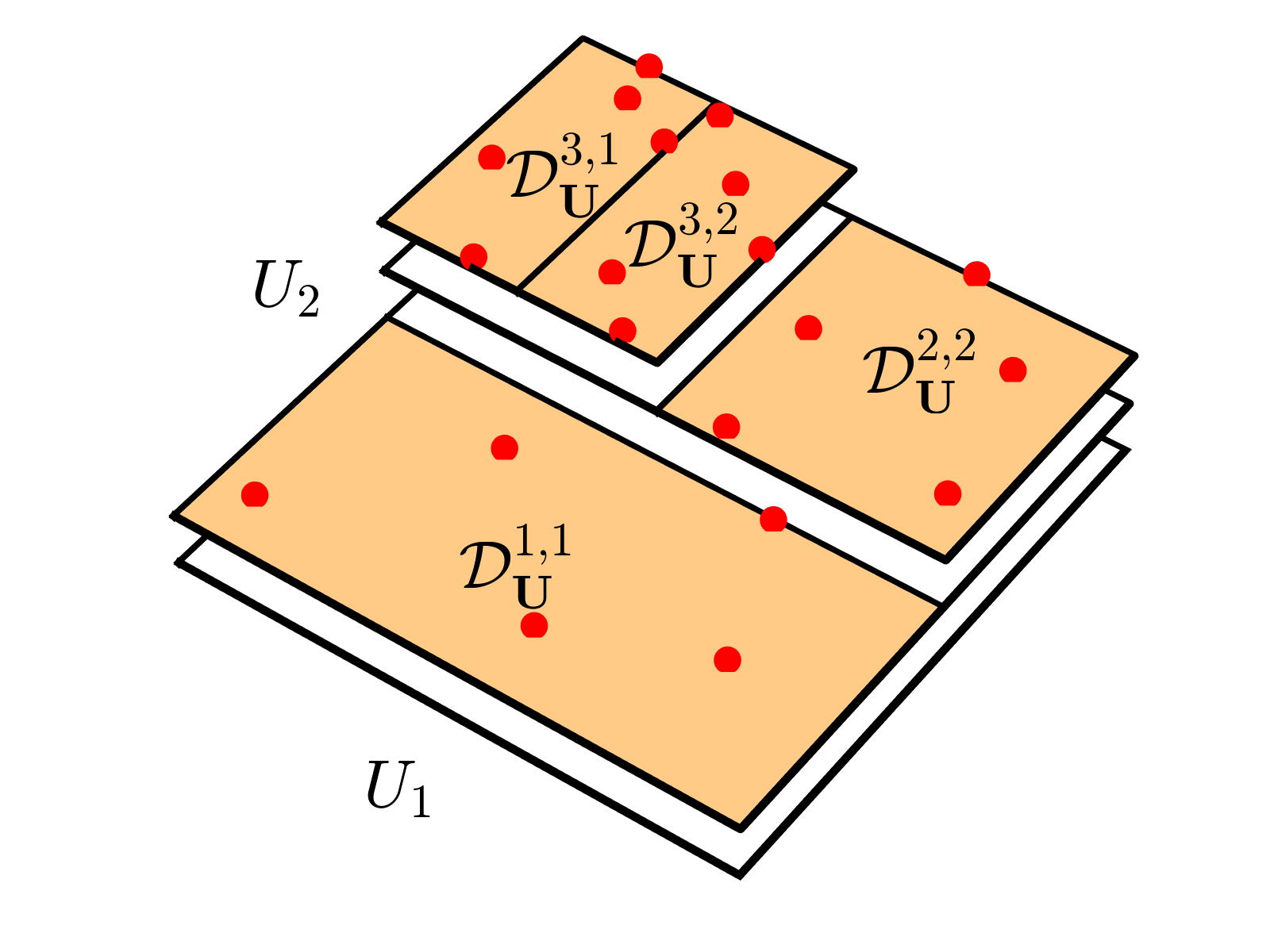}
		\end{minipage}
	}%
	\subfloat[Real space]{
		\label{fig:cs1:SSEBehaviour:2}
		\begin{minipage}{0.5\linewidth}
			\centering
			\includegraphics[width=0.7\linewidth, trim = 0 10 0 12 clip]{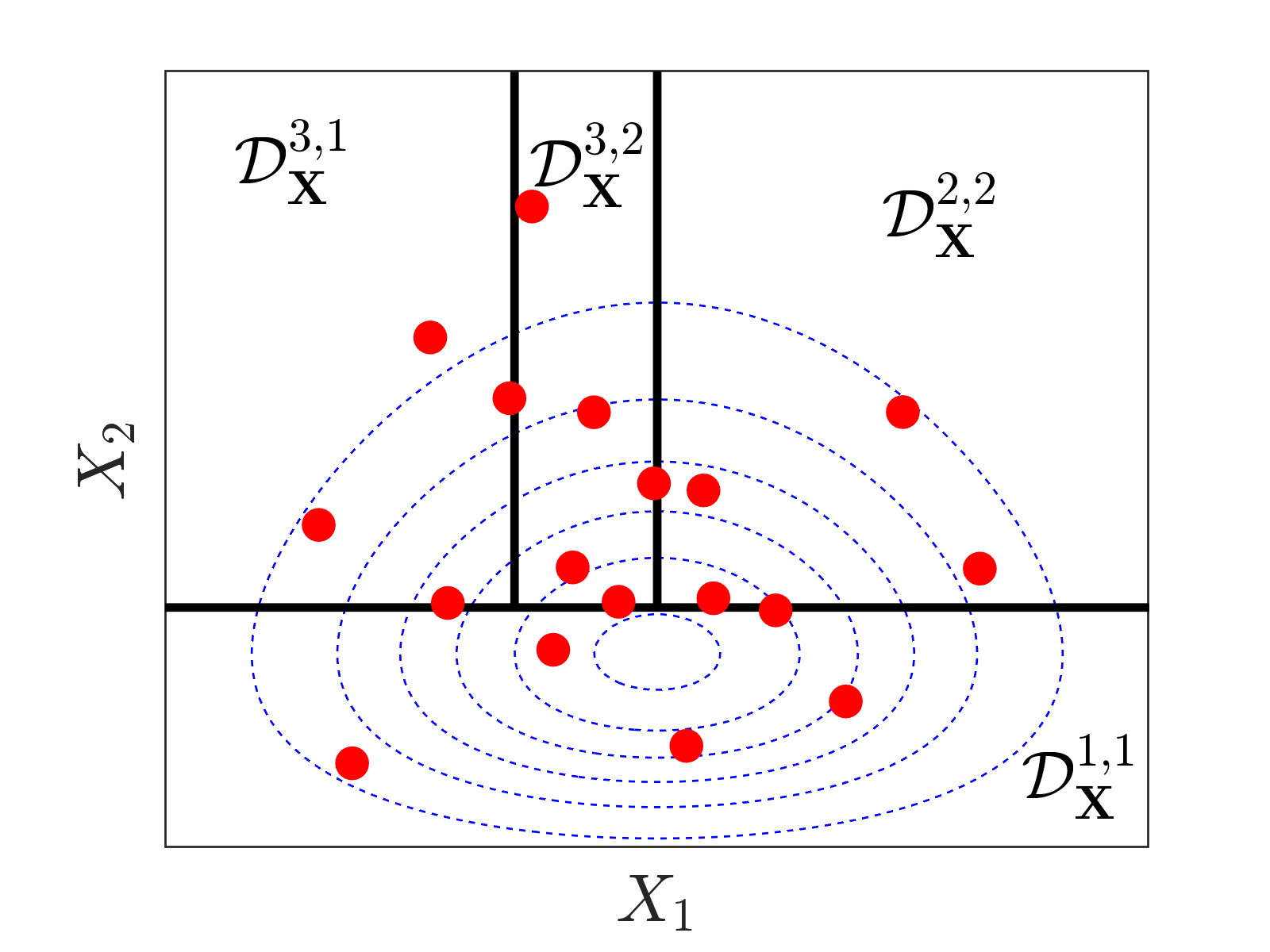}
		\end{minipage}
	}%
	\caption{\emph{Stochastic spectral embedding}: Illustration of SSE representation in Eq.~\eqref{eq:SSErepres} in the quantile and real space. Red points denote the experimental design and orange areas denote terminal or unsplit domains, that partition the input space into a set of \emph{disjoint} subdomains such that $\cD_{\vX} = \bigcup_{k\in\cT} \cD_{\vX}^k$.}%
	\label{fig:SSErepres}%
\end{figure}

Stochastic spectral embedding (SSE, \cite{Marelli2020}) is a function approximation technique that works by sequentially expanding residuals $\widehat{\cR}_S^{k}(\vX)$ in subdomains of the parameter space $\cD_{\vX}^k\subseteq\cD_{\ve{X}}$. At each \rev{level} in the sequence, SSE constructs local spectral expansions of the residual of the current approximation at a \rev{set of} subdomains, as illustrated in Figure~\ref{fig:SSErepres}. As for any other metamodeling techniques, this approach can be used in reliability analyses to directly approximate the limit-state function:
\begin{equation}
	\label{eq:SSErepres}
	g(\ve{X}) \approx g_{\text{SSE}}(\ve{X}) = \sum\limits_{k\in\cK} \indfun{\cD_{\vX}^{k}}(\vX)\, \widehat{\cR}_S^{k}(\vX). 
\end{equation}
\rev{In this expression, $\cK\subseteq\mathbb{N}^2$ is a set of index pairs with elements $k=(\ell,p)$ used to identify the residual expansions in the SSE representation. The first element of the index pair $\ell \in \{0,\cdots,L\}$ denotes the level at which the residual expansion resides, where $L$ is the total number of levels. The second element $p \in \{1,\cdots,P_\ell\}$ denotes the index of the residual expansion at a given level, with $P_\ell$ denoting the total number of expansions at the $\ell$-th level.} $\indfun{\cD_{\vX}^{k}}$ is the indicator function that is equal to $1$ in the $k$-th subdomain and $0$ everywhere else. For further derivations, it is beneficial to distinguish between standard and \emph{terminal domains} among the $\card{\cK}$ domains used in the SSE representation. \rev{A terminal domain is any domain that has not been split. It can be located at any level of the SSE representation and the set of all terminal domains with indices gathered in $\cT$ constitutes a complete partition of the input domain, \ie $\cD_{\vX}=\bigcup_{k\in\cT}\cD_{\vX}^k$ (top domains highlighted in orange in Figure~\subref{fig:cs1:SSEBehaviour:1}).} With this, Eq.~\eqref{eq:SSErepres} is expanded to
\begin{equation}
	\label{eq:SSErepresExtend}
	g_{\text{SSE}}(\ve{X}) = \sum\limits_{k\in\cK\setminus\cT} \indfun{\cD_{\vX}^{k}}(\vX)\, \widehat{\cR}_S^{k}(\vX) + \sum\limits_{k\in\cT} \indfun{\cD_{\vX}^{k}}(\vX)\, \widehat{\cR}_S^{k}(\vX), 
\end{equation}

In principle, any spectral expansion technique could be used to expand the residuals \citep{Marelli2020}. In the present contribution, we consider only \emph{polynomial chaos expansions} (PCEs) such that
\begin{equation}
	\label{eq:resExp}
	\widehat{\cR}_S^{k}(\vX) \eqdef \sum_{\ve{\alpha}\in\cA^k}a^k_{\ve{\alpha}}\Psi^k_{\ve{\alpha}}(\vX),
\end{equation}
where $\mathcal{A}^k$ is a multi-index set describing the dimension-wise polynomial degrees, $\{\Psi_{\ve{\alpha}}^k\}_{\ve{\alpha}\in\cA^k}$ are mutually orthogonal polynomial basis functions and $a_{\ve{\alpha}}^k$ denote the corresponding coefficients \citep{Xiu2002, Blatman2011a, Luethen2020sparse}. 

For computational and bookkeeping ease, we manage the SSE construction in what we call the \emph{quantile space}, \ie the $M$-dimensional unit hypercube denoted by $\cD_{\ve{U}}$. As discussed in \citet{Marelli2020}, for every original or \emph{real input space} $\cD_{\vX}$ there exists an isoprobabilistic mapping to $\cD_{\ve{U}}$ (Figure~\subref{fig:cs1:SSEBehaviour:1}), such that the partitioning of the input space can be conducted in the quantile space without any loss of generality.

The SSE representation shown in Figure~\ref{fig:SSErepres} ultimately partitions the input space into $\card{\cT}$ disjoint subdomains such that $\bigcup_{k\in\cT}\cD_{\vX}^k=\cD_{\vX}$. Using this representation, the failure probability from Eq.~\eqref{eq:failureProb} can be rewritten as a sum of \emph{\rev{conditional} failure probabilities}
\begin{equation}
	\label{eq:failureProb_SSE}
	P_f = \sum_{k\in\cT} \cV^{k} P_f^k, \quad \text{with} \quad \cV^{k} = \int\limits_{\cD_{\vX}^{k}} f_{\vX}(\vx)\,\di\vx,
\end{equation} 
where $\cV^k$ is the \emph{domain-wise probability mass} that is readily available from the construction in the quantile space, and $P_f^k$ is the $k$-th \rev{conditional} failure probability given by
\begin{equation}
	\label{eq:failureProb_SSE_local}
	P_f^k\eqdef\mathbb{P}\left[g(\ve{X})\le 0\vert\ve{X}\in\cD_{\vX}^k\right] \approx \mathbb{P}\left[g_{\text{SSE}}(\ve{X})\le 0\vert\ve{X}\in\cD_{\vX}^k\right].
\end{equation}
This \rev{conditional} failure probability is estimated here with suitable probabilistic simulation methods utilizing the SSE representation of the limit-state function (more details are provided in Section~\ref{sec:estimatingLocFail}). Assuming the simulation method is accurate enough, the accuracy of the failure probability estimate depends only on that of the SSE representation itself. 

To devise an adaptive algorithm, it is necessary to devise a suitable error measure on each \rev{conditional} failure probability estimate. As shown in \citet{Marelli2020}, the SSE representation provides \emph{domain-wise} prediction error estimators, \eg by means of cross-validation (\eg leave-one-out error). Such estimators can in principle be used as a proxy to assess the accuracy of $P_f^k$. 
However, these estimators are not robust in the presence of very localized behaviour that is typical for limit-state functions. An improved estimator that takes into account the \emph{point-wise} prediction accuracy can be derived by means of the local error estimation of bootstrap PCE (bPCE), originally introduced in a reliability context in \citet{MarelliSS2018}.

\subsubsection{Bootstrap SSE}
The idea of bPCE is to use bootstrap resampling \citep{Efron1979} on the original experimental design to create a set of $B$ experimental designs and construct $B$ sparse PCEs with them. The individual bootstrap PCE realizations can then be used to estimate point-wise statistics of the PCE prediction.

This idea can be applied to the residual expansions within the SSE approach. Due to the sequential embedding of the residual expansions, the prediction error of the full SSE approximation can be assessed solely based on the prediction errors in the terminal domains \citep{Marelli2020}. In every terminal domain, we therefore construct $B$ residual expansions to obtain $B$ bootstrap predictions
\begin{equation}
	\label{eq:SSEbootstrap}
	g_{\text{SSE}}^{(b)}(\ve{X}) = \sum\limits_{k\in\cK\setminus\cT} \indfun{\cD_{\vX}^{k}}(\vX)\, \widehat{\cR}_S^{k}(\vX) + \sum\limits_{k\in\cT} \indfun{\cD_{\vX}^{k}}(\vX)\, \widehat{\cR}_S^{k,(b)}(\vX), \quad \text{for} \quad b \in \{1,\cdots, B\}.
\end{equation}

By analogy with \citet{MarelliSS2018}, we can define an estimate of the point-wise misclassification probability as:
\begin{equation}
	\label{eq:misclassificationBootstrap}
	\hat{P}_{m}(\vx)\eqdef\frac{1}{B}\sum_{b=1}^B\left\vert\indfun{\cD_{f,\mathrm{SSE}}}(\vx)-\indfun{\cD_{f,\mathrm{SSE}}}^{(b)}(\vx)\right\vert,
\end{equation}
where $\indfun{\cD_{f,\mathrm{SSE}}}$ and $\indfun{\cD_{f,\mathrm{SSE}}}^{(b)}$ are evaluated with the mean and bootstrap predictors of SSE respectively (see also Eq.~\eqref{eq:failureInd}).

The bootstrap replications can also be used to directly obtain statistics of the \rev{conditional} failure probability estimates, such as variance or confidence bounds. As an example, the \rev{conditional} failure probability variance is given by
\begin{equation}
	\label{eq:failureProb_SSE_VarianceDomain}
	\Var{\hat{P}_f^k} = \Var{\left\{\hat{P}_f^{k,(1)},\cdots,\hat{P}_f^{k,(B)}\right\}},
\end{equation}
with the failure probability of the $b$-th replication given by
\begin{equation}
	\label{eq:failureProb_SSE_localBootstrap}
	\hat{P}_f^{k,(b)} \eqdef \mathbb{P}\left[g_{\text{SSE}}^{(b)}(\ve{X})\le 0\vert\ve{X}\in\cD_{\vX}^k\right].
\end{equation}

By means of Eq.~\eqref{eq:failureProb_SSE} this \rev{conditional} failure probability variance can be used to write an expression for the total failure probability variance as
\begin{equation}
	\label{eq:failureProb_SSE_Variance}
	\Var{\hat{P}_f} = \sum_{k\in\cT}\left(\cV^k\right)^2\Var{\hat{P}_f^k}.
\end{equation}
\rev{This expression assumes independence between the conditional failure probability estimates. Due to the ancestry relation between subdomains, we cannot in general rule out dependence between those quantities, however, we assume that it is negligible because all expansions in terminal domains are constructed with independently enriched samples.}

Similarly, the bootstrap replications can be used to define \emph{confidence bounds} on the total failure probability. Let $\hat{P}_f^{(b)}\eqdef\sum_{k\in\cT}\cV^k \hat{P}_f^{k,(b)}$ be the total failure probability estimated with the $b$-th replication from all terminal domains. The equal tail confidence bounds $\left\{\overline{\hat{P}_f^k}, \underline{\hat{P}_f^k}\right\}$ are then defined by
\begin{equation}
	\label{eq:failureBounds}
	\begin{split}
		\mathbb{P}\left[\underline{\hat{P}_f^k} \le \left\{\hat{P}_f^{(1)},\cdots,\hat{P}_f^{(B)}\right\}\right] &\approx \alpha,\\
		\mathbb{P}\left[\left\{\hat{P}_f^{(1)},\cdots,\hat{P}_f^{(B)}\right\} \le \overline{\hat{P}_f^k}\right] &\approx 1-\alpha\\
	\end{split}
\end{equation}
with $\alpha\in[0,0.5]$ and $\gamma\eqdef1-2\alpha$ is called the \emph{symmetrical confidence level}. It is common practice to take $\gamma=95\%$ and $\alpha=2.5\%$.

\subsubsection{Dependent input parameters}
Generally, the input random vector $\ve{X}$ may have mutually dependent marginals that we treat with copula theory in this work \citep{Nelsen2006, Joe2015}. A typical approach to address dependence in reliability problems is to use isoprobabilistic mappings to an independent space, \eg the standard normal space. However, sparse spectral techniques acquire their strength from the \emph{sparsity of effects principle} \citep{Montgomery2019}, which states that in many engineering models the majority of the output is attributable to low-interaction order terms. A mapping to the standard normal space might therefore reduce the efficiency of those techniques.

Nonetheless, dependence is challenging for the construction of the spectral basis functions $\Psi_{\ve{\alpha}}^k$ in Eq.~\eqref{eq:resExp}. However, as extensively studied in \citet{Torre2019PCE4ML}, polynomial chaos expansions seem to be most accurate for predictive purposes when dependence is simply ignored and the basis is derived by tensor product of univariate bases orthogonal to the marginals instead. Applying this to SSE suggests it is best to construct the domain-wise polynomial basis by assuming the domain-wise marginals to be independent. 

\begin{figure}
	\centering
	\subfloat[Quantile space]{
		\label{fig:dependence:quantile}
		\begin{minipage}{0.5\linewidth}
			\centering
			\includegraphics[width=\linewidth, trim = 0 10 0 12]{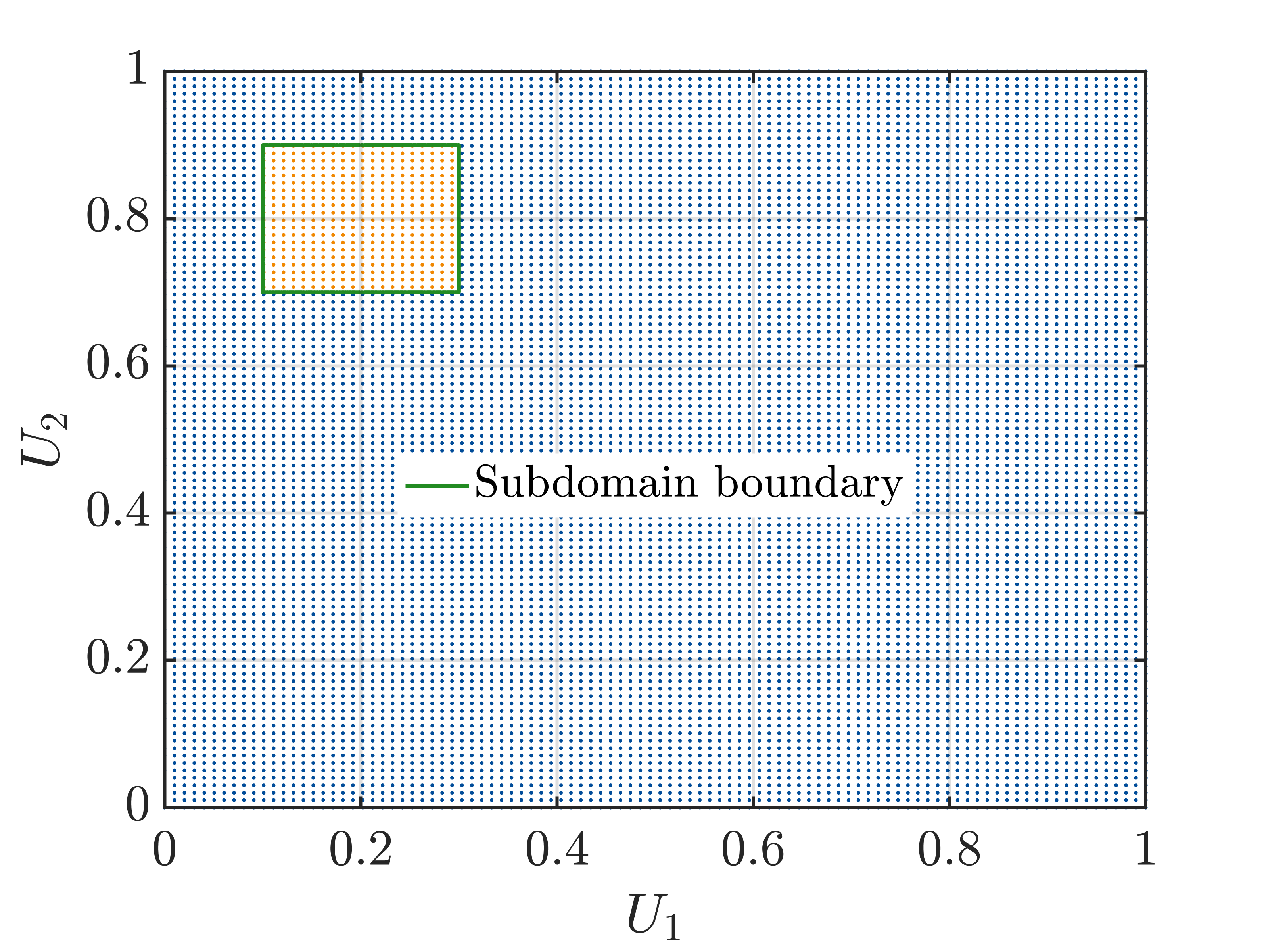}
		\end{minipage}
	}%
	\subfloat[Real space]{
		\label{fig:dependence:real}
		\begin{minipage}{0.5\linewidth}
			\centering
			\includegraphics[width=\linewidth, trim = 0 10 0 12]{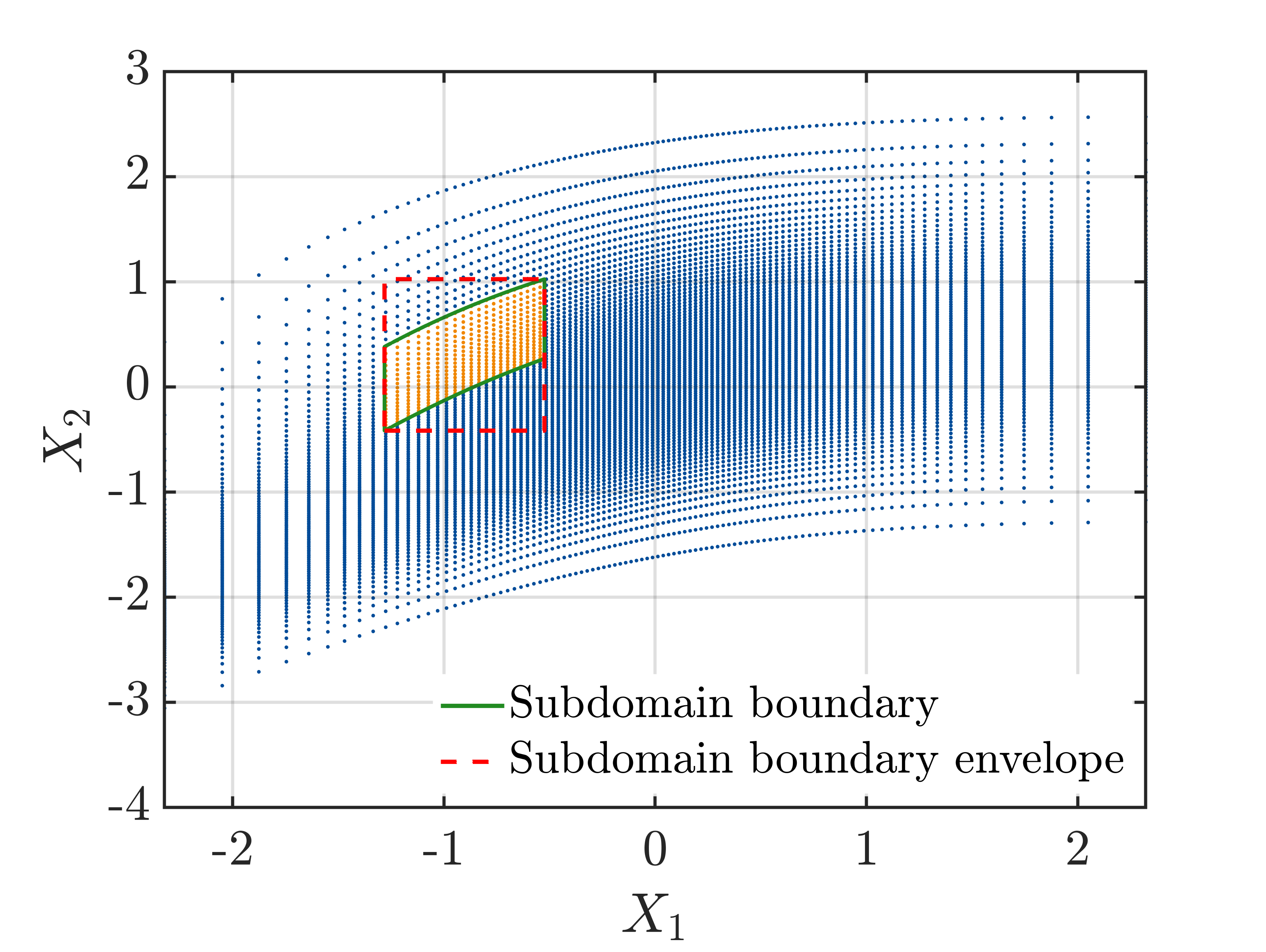}
		\end{minipage}
	}%
	\caption{\emph{Stochastic spectral embedding}: Illustration of how dependence is handled in the algorithm. For two mutually dependent input parameters $X_1$ and $X_2$, the plots show a set of sample points in the real and quantile space with a subset of points inside an exemplary subdomain. The parameter dependence leads to a loss of orthogonality of the subdomains in the real space, which necessitates the introduction of an \emph{enveloping hypercube} in the real space to construct the spectral basis functions $\Psi_{\ve{\alpha}}^k$.}%
	\label{fig:dependence}%
\end{figure}

As mentioned before, the partitioning into orthogonal subdomains is defined in the independent quantile space \citep{Marelli2020}. Therefore, the transformation of those domains to the dependent physical space results in non-orthogonal domains (see Figure~\ref{fig:dependence}). Constructing the univariate polynomial bases requires knowledge of the marginal bounds in the physical space, which cannot be computed in a straightforward fashion.

To approximate these bounds, we therefore (1) randomly sample the $(M-1)$-dimensional boundary of the hypercube in the quantile space, (2) transform those points to the real space using a suitable isoprobabilistic transform \citep{Rosenblatt1952, Torre2019PEM} and proceed to (3) compute the basis with an orthogonal \emph{enveloping hypercube} in the physical space around those transformed points (see Figure~\subref{fig:dependence:real}). 
It is interesting to note that for prediction purposes, the resulting multivariate basis will only be evaluated inside the \emph{subdomain boundary} and not the \emph{subdomain boundary envelope}, as the region outside the subdomain boundary belongs to neighbouring subdomains.

\subsection{An updated sequential partitioning algorithm}
\label{sec:changesAlgo}

Failure domains typically occupy only a small fraction of the input space. The original \emph{equal probability} partitioning strategy presented in \citet{Marelli2020, Wagner2020JCP} is therefore expected to converge only slowly to those domains. We therefore propose modifications that make the algorithm more efficient for reliability problems. The structure of the proposed algorithm follows the sequential partitioning algorithm used in the adaptive SSLE approach presented in \citet{Wagner2020JCP}. It is repeated here in an abbreviated form. 

\begin{enumerate}\itemsep0pt
	\item \textbf{Initialization}
	\begin{enumerate}
		\item Sample from input distribution
		\item Calculate initial expansion
	\end{enumerate} 
	\item \label{algoAdaptAbbrev:RefDomSelection} \textbf{Select refinement domain}:
	\begin{enumerate}\itemsep0pt\small
		\item \label{algoAdaptAbbrev:Partition} Partition domain
		\item \textbf{For each subdomain}:
		\begin{enumerate}
			\item \label{algoAdaptAbbrev:Enrich} Enrich sample
			\item Create residual expansion
			\item Update domain-wise error
		\end{enumerate}
		\item Go to Step~\ref{algoAdaptAbbrev:RefDomSelection}
	\end{enumerate}
\end{enumerate} 

The main modifications to the algorithm that are specific to reliability problems pertain to the \emph{refinement domain selection} (Step~\ref{algoAdaptAbbrev:RefDomSelection}), \emph{partitioning} (Step~\ref{algoAdaptAbbrev:Partition}) and \emph{sample enrichment} (Step~\ref{algoAdaptAbbrev:Enrich}) strategies, which utilize the point-wise prediction error estimators available from bPCE. These modifications are detailed in the following sections. The full SSE-based reliability (SSER) algorithm is sketched in Section~\ref{sec:changesAlgo:algorithm}.

\subsubsection{Refinement domain selection}
\label{sec:changesAlgo:refDom}
At every step, the adaptive sequential partitioning algorithm chooses a \emph{refinement domain} from the set of terminal domains $\cT$ to be split and enriched with sample points. To choose a domain, the terminal domains are ranked based on their importance to the overall approximation. In the present setting, the goal of refinement is the reduction of uncertainty associated with the failure probability. Viewing Eq.~\eqref{eq:failureProb_SSE_Variance} as a variance decomposition of the estimator variance, it is natural to refine the terminal domain that has the largest contribution to this variance estimator. More formally, out of the terminal domains we refine the domain with the largest
\begin{equation}
	\label{eq:algo:error}
	\cE^{k} \eqdef \left(\cV^{k}\right)^2\Var{\hat{P}_f^{k}},
\end{equation} 
%

This equation depends on the domain-wise probability mass $\cV^k$ and the variance of the \rev{conditional} failure probability estimator $\Var{\hat{P}_f^{k}}$. In practice, $\Var{\hat{P}_f^{k}}$ is estimated with Eq.~\eqref{eq:failureProb_SSE_VarianceDomain}, which in turn depends on the $B$ estimates of $\hat{P}_f^{k,(b)}$. These estimates are computed with accurate simulation methods using the bootstrap SSE approximations (see Section~\ref{sec:estimatingLocFail}). 

If the SSE approximation in the current subdomain is not sufficiently accurate, it can happen that all bootstrap replications miss existing failure regions, ultimately resulting in a gross underestimation of $\cE^{k}$. To avoid overlooking such domains as possible refinement domains, the algorithm additionally prioritizes them based on the probability mass $\cV^{k}$. The refinement domain is ultimately chosen as 
\begin{equation}
	\label{eq:algo:refinement}
	k_{\mathrm{refine}, i} = \argmax_{k\in\cT}
	\begin{cases}
		\cV^{k}, \quad & \text{if Eq.~\eqref{eq:interStoppingCriterion} is met},\\
		\cE^{k}, \quad & \text{otherwise},
	\end{cases}
\end{equation}
which depends on the following \emph{intermediate re-prioritization criterion} 
\begin{equation}
	\label{eq:interStoppingCriterion}
	\frac{\Var{\rev{\big\{}\hat{P}_f^{(i-2)},\hat{P}_f^{(i-1)},\hat{P}_f^{(i)}\rev{\big\}}}}{\left(\hat{P}_f^{(i)}\right)^2} < \varepsilon_{\hat{P}_f},
\end{equation}
where $\hat{P}_f^{(i)}$ is the total failure probability estimator at the $i$-th iteration of the algorithm and the threshold is heuristically chosen to $\varepsilon_{\hat{P}_f}=0.1\%$. This criterion is triggered when the failure probability estimator does not change significantly in three successive iterations.

\subsubsection{Partitioning strategy}
\label{sec:changesAlgo:partition}

\begin{figure}
	\centering
	\includegraphics[width=0.8\linewidth]{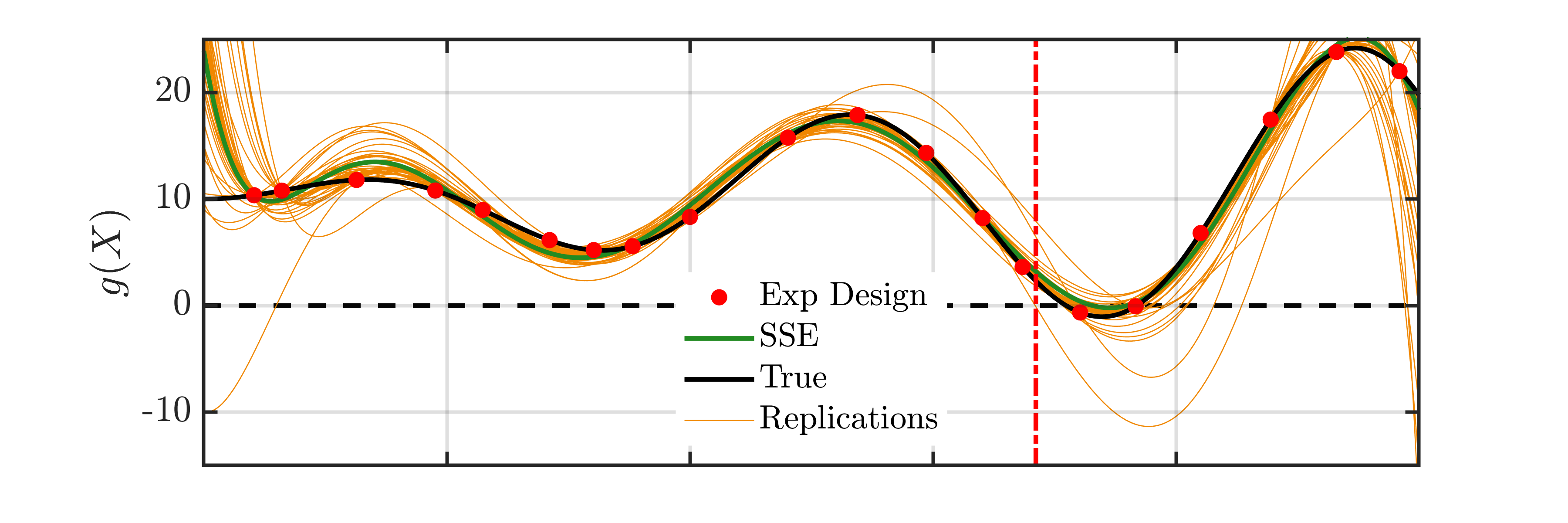}\\
	\includegraphics[width=0.8\linewidth]{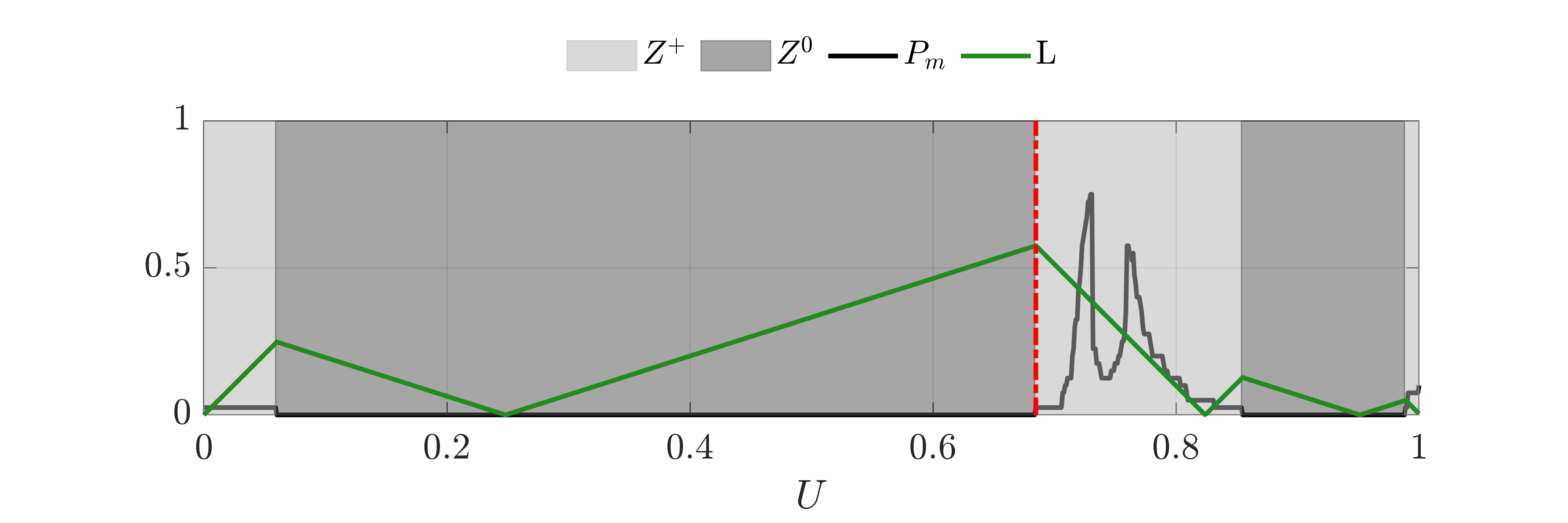}
	\caption{Partitioning strategy exemplified on a one dimensional function showing the bootstrap replications and the resulting support of $Z^+$ and $Z^0$ as well as the objective function $L$. The splitting location $\upsilon$ is highlighted as a vertical, dashed red line. The plots are shown in the quantile space $\cD_U$. \label{fig:Method:partitioning}}
\end{figure}

Once a refinement domain $\cD_{\vX}^{k_{\mathrm{refine}}}$ is selected, the partitioning strategy determines how it is split. Refinement domains are split into two subdomains with unequal probability mass. For notational simplicity, we omit the superscript $k_{\mathrm{refine}}$ denoting the refinement domain in this section and emphasize instead that the following equations are valid for all $k\in\cT$. For illustrative purposes, the partitioning strategy is exemplified for a simple problem in Figure~\ref{fig:Method:partitioning}.

To maximize enrichment efficiency, we base our splitting strategy on separating regions that predict failure/safety correctly from those that do so incorrectly. A measure of prediction accuracy in this respect is given by the misclassification probability $P_m$ \citep{Echard2011} defined for bootstrap SSE in Eq.~\eqref{eq:misclassificationBootstrap}. To this end, we define two auxiliary conditional random vectors in the quantile space $\cD_{\ve{U}}$ as
\begin{equation}
	\label{eq:auxVariable}
	\vZ^{0}\eqdef \ve{U}\vert \hat{P}_{m}=0,\quad \text{and} \quad \vZ^{+}\eqdef \ve{U}\vert \hat{P}_{m}>0.
\end{equation}
These vectors identify regions of zero ($\vZ^{0}$) and non-zero ($\vZ^{+}$) misclassification probability in the current refinement domain.



Based on these random vectors, we compute the location of the split. We choose a set of \emph{splitting locations} in each dimension $\upsilon_i\in \cD_{U_{i}},i=1,\cdots,M$ that ideally completely separate the support of $\vZ^+$ from $\vZ^0$ (see Figure~\ref{fig:Method:partitioning}). Due to the likely disjoint support of each auxiliary random vector, the split should instead confine a maximum of $\vZ^+$'s probability mass to one side of the split, while confining a maximum of $\vZ^0$'s probability mass to the other side of the split. Denoting by $Z^+_i$ and $Z^0_i$ the marginals of the auxiliary random vectors in the $i$-th dimension, we pick the splitting location
\begin{equation}
	\label{eq:partStrategy}
	\widehat{\upsilon}_i = \arg\max_{\upsilon_i\in \cD_{X_{i}}}L_i(\upsilon_i), \quad \text{with} \quad
	L_i(\upsilon_i) \eqdef -1 + \max
	\begin{cases}
		\mathbb{P}\left[Z_i^+\le\upsilon_i\right] + \mathbb{P}\left[Z_i^0>\upsilon_i\right],\\
		\mathbb{P}\left[Z_i^+>\upsilon_i\right] + \mathbb{P}\left[Z_i^0\le\upsilon_i\right],
	\end{cases}
\end{equation}
where the objective function $L_i$ characterizes the split properties by returning the maximum of the respective auxiliary probability masses in the resulting subdomains. The splitting location $\widehat{\upsilon}_i$ that maximizes $L_i$ thus splits the initial domain $\cD_{\vX}$ into two subdomains $\cD_{X_i}^{(1)}$ and $\cD_{X_i}^{(2)}$ that fulfil the initially stated goal of optimally separating regions that correctly predict failure/safety from regions that do so incorrectly. The objective function $L_i$ is bounded between $[0, 1]$, where a value of $1$ indicates a \emph{perfect split} that creates one subdomain with only non-zero and one subdomain with all zero misclassification probability regions. 


To ultimately choose a \emph{splitting direction} $d\in\{1,\cdots,M\}$, we compare the values of the objective functions $L_i(\widehat{\upsilon}_i)$ and split along the dimension that achieves the best split, \ie	
\begin{equation}
	\label{eq:splitDirection}
	d = \arg\max_{i\in \{1,\cdots,M\}} L_i(\widehat{\upsilon}_i).
\end{equation}


In practice, a sufficiently large sample of the auxiliary random vectors $\vZ^0$ and $\vZ^+$ is used to conduct the computations of this section. This sample is readily available, as the SSE surrogate model can be evaluated at a negligible computational cost.

\paragraph{No misclassified sample points}\mbox{}\\
It might occur that the the algorithm does not detect misclassified sample points at the time of partitioning. When this happens, the misclassification probability $P_m$ (Eq.~\eqref{eq:misclassificationBootstrap}) is zero everywhere and consequently the auxiliary random vector $\vZ^+$ is not defined. Because of the preceding refinement selection criterion in Eq.~\eqref{eq:algo:refinement} this can only happen at the first step of the algorithm or after the intermediate re-prioritization criterion (Eq.~\eqref{eq:interStoppingCriterion}) has been triggered.

To let the algorithm proceed in this case, the definition of the auxiliary random vectors is updated to discriminate based on the probability to predict values within a predefined empirical quantile around the limit-state surface. More formally, $P_m$ in Eq.~\eqref{eq:auxVariable} is then replaced with 
\begin{equation}
	P_t(\vx) \eqdef\frac{1}{B}\sum_{b=1}^B\indfun{\cD_t}^{(b)}(\vx),
\end{equation}
where $\indfun{\cD_t}^{(b)}$ is defined with an empirical quantile $t$ such that $\mathbb{P}\left[\vert g_{\mathrm{SSE}}(\ve{X})\vert\le t\right] = \varepsilon_{t}$ as
\begin{equation}
	\indfun{\cD_t}^{(b)}(\ve{x}) \eqdef
	\begin{cases}
		1, \quad \text{if} \quad \vert g_{\mathrm{SSE}}^{(b)}(\ve{x})\vert \le t ,\\
		0, \quad \text{if} \quad \vert g_{\mathrm{SSE}}^{(b)}(\ve{x})\vert > t.
	\end{cases}
\end{equation}
The value $\varepsilon_{t}=0.01$ has proven to be a good choice in practice.

\subsubsection{Sample enrichment}
\label{sec:changesAlgo:sampleEnrichment}
After partitioning a selected refinement domain, in the original SSE algorithm points are added such that every subdomain contains exactly a total of $\NRefine$ points. Because the partitioning strategy now allows the creation of domains with considerably different probability mass, the original sample enrichment approach is problematic. As an example, splitting in \rev{the tails of the input distribution} will result in one large domain with $\NRefine$ points from the previous step and one smaller domain with $0$ points. In this case, no new points would be added to the large domain. 
The subsequently constructed residual expansion would be based on (almost) the same information as the parent domain expansion and be prone to over-fitting. We therefore employ a different strategy here, which always adds $\NRefine$ points to every domain, independent of the existing points inside that domain. 
We refer to those points as the \emph{sample budget} here.

The second change pertains to the placement of the sample budget. The original algorithm used random sampling to place points uniformly in the refinement subdomain. To exploit the fact that the SSE for reliability applications needs to be more accurate near the limit-state surface, we propose a slightly modified strategy: we split the sample budget and randomly place 
\begin{equation}
	N_{\mathrm{uni}} = \max
	\begin{cases}
		\frac{\NRefine}{2}-\frac{N_{\mathrm{curr}}}{2}\\
		0
	\end{cases}
\end{equation}
points uniformly in the refinement domain quantile space, where $N_{\mathrm{curr}}$ is the number of existing points in the refinement domain. The remainder of the sample budget is used to randomly place sample points in the subset of the refinement domain with non-zero misclassification probability $P_m$. This corresponds to sampling the auxiliary random vector $\vZ^+$ defined in Eq.~\eqref{eq:auxVariable}. We do this to ensure stability of the residual expansions that are known to deteriorate in accuracy when the experimental design clusters in a subspace.

In the extreme case of no existing points (\ie $N_{\mathrm{curr}}=0$), this places half of the sample budget uniformly and the other half according to $\vZ^+$. In the other extreme case of $N_{\mathrm{curr}}\ge\NRefine$, the whole sample budget is devoted to sampling $\vZ^+$. 

At the first iteration, to construct the initial expansion, the enrichment occurs randomly in the entire input quantile space. We observed from repeated tests that the algorithm is less stable if the initial expansion is less accurate. To account for this, we start out with an initial experimental design of size $2\NRefine$ and switch to the experimental design size of $\NRefine$ in all subsequent iterations.

Finally, we note that $\NRefine$ is a tuning parameter of the algorithm. Larger values of $\NRefine$ lead to more accurate local expansions, resulting in increased stability of the algorithm at the cost of slower convergence rates. Vice-versa, smaller values of $\NRefine$ can significantly increase the convergence rates, but can potentially produce inaccurate limit-state function approximations that lead to biased failure probability estimates. 

\subsubsection{Stochastic spectral embedding-based reliability algorithm}
\label{sec:changesAlgo:algorithm}

Taking all presented ingredients, the full stochastic spectral embedding-based reliability (SSER) algorithm is written as follows:

\begin{enumerate}\itemsep0pt
	\item[] \textbf{Input:}
	\begin{itemize}\itemsep0pt\small
		\item Input parameters $\vX$, limit-state function $g$
		\item Expansion parameters $\NRefine, p_{\mathrm{max}}, B$
		\item Algorithm parameters $\Ntot, \varepsilon_{\hat{\beta}}, \varepsilon_{\hat{P}_f}, \varepsilon_{t}$
	\end{itemize}
	\item \textbf{Initialization:}
	\begin{enumerate}\itemsep0pt\small
		\item $\cD_{\vX}^{0,1} = \cD_{\vX}$
		\item Sample from the input distribution an initial experimental design $\cX = \{\vx^{(1)},\cdots,\vx^{(2\NRefine)}\}$
		\item Calculate the truncated expansion $\widehat{\cR}_S^{0,1}(\vX)$ of $g(\vX)$ in the full domain $\cD_{\vX}^{0,1}$, retrieve its approximation error $\cE^{0,1}$ and initialize $\cT = \{(0,1)\}$
		\item ${\cR}^{1}(\vX) = g(\vX) - \widehat{\cR}_S^{0,1}(\vX)$
	\end{enumerate}
	\item \label{algoAdapt:2} \textbf{For $(\ell,p) = k_{\mathrm{refine}}$ from Eq.~\eqref{eq:algo:refinement}}
	\begin{enumerate}\itemsep0pt\small
		\item \label{algoAdapt:updateT} Split the current subdomain $\cD_{\vX}^{\ell,p}$ in $2$ sub-parts $\cD_{\vX}^{\ell+1,\{s_1,s_2\}}$ according to Eq.~\eqref{eq:splitDirection}\rev{, where $s_1$ and $s_2$ are unique indices for domains on the $\ell$-th level}, and update $\cT$
		\item \textbf{For each split $s \rev{\in} \{s_1, s_2\}$}
		\begin{enumerate}
			\item \label{algoAdapt:enrichED} Enrich sample $\cX$ with $\min\{\NRefine,\Ntot-\vert\cX\vert\}$ new points inside $\cD_{\vX}^{\ell+1,s}$
			\item \textbf{If} $\NRefine$ new points inside $\cX$
			\begin{enumerate}
				\item \label{algoAdapt:createResExp} Create the truncated expansion $\widehat{\cR}_S^{\ell+1,s}(\vX)$ of the residual ${\cR}^{\ell+1}(\vX)$ in the current subdomain using a subset of $\cX$ inside $\cD_{\vX}^{\ell+1,s}$
				\item Update the residual on the next level in the current subdomain ${\cR}^{\ell+2}(\vX) = {\cR}^{\ell+1}(\vX) - \widehat{\cR}_S^{\ell+1,s}(\vX)$
			\end{enumerate}
			\item \label{algoAdapt:Error} Retrieve $\cE^{\ell+1,s}$ from Eq.~\eqref{eq:algo:error}
		\end{enumerate}
		\item Retrieve $\hat{P}_f$ from Eq.~\eqref{eq:failureProb_SSE} and its variance from Eq.~\eqref{eq:failureProb_SSE_Variance} 
		\item \textbf{If} stopping criterion in Eq.~\eqref{eq:stoppingCriterion} is met or less than two new expansions were created, terminate the algorithm, otherwise go back to \ref{algoAdapt:2}
	\end{enumerate}
	\item \textbf{Termination}
	\begin{enumerate}\itemsep0pt\small
		\item Return $\hat{P}_f$ and its variance
	\end{enumerate}
\end{enumerate}

\begin{figure}
	\centering
	\subfloat[Initialization]{
		\begin{minipage}{0.33\linewidth}
			\includegraphics[width=\linewidth,clip=true,trim=0 0 0 0]{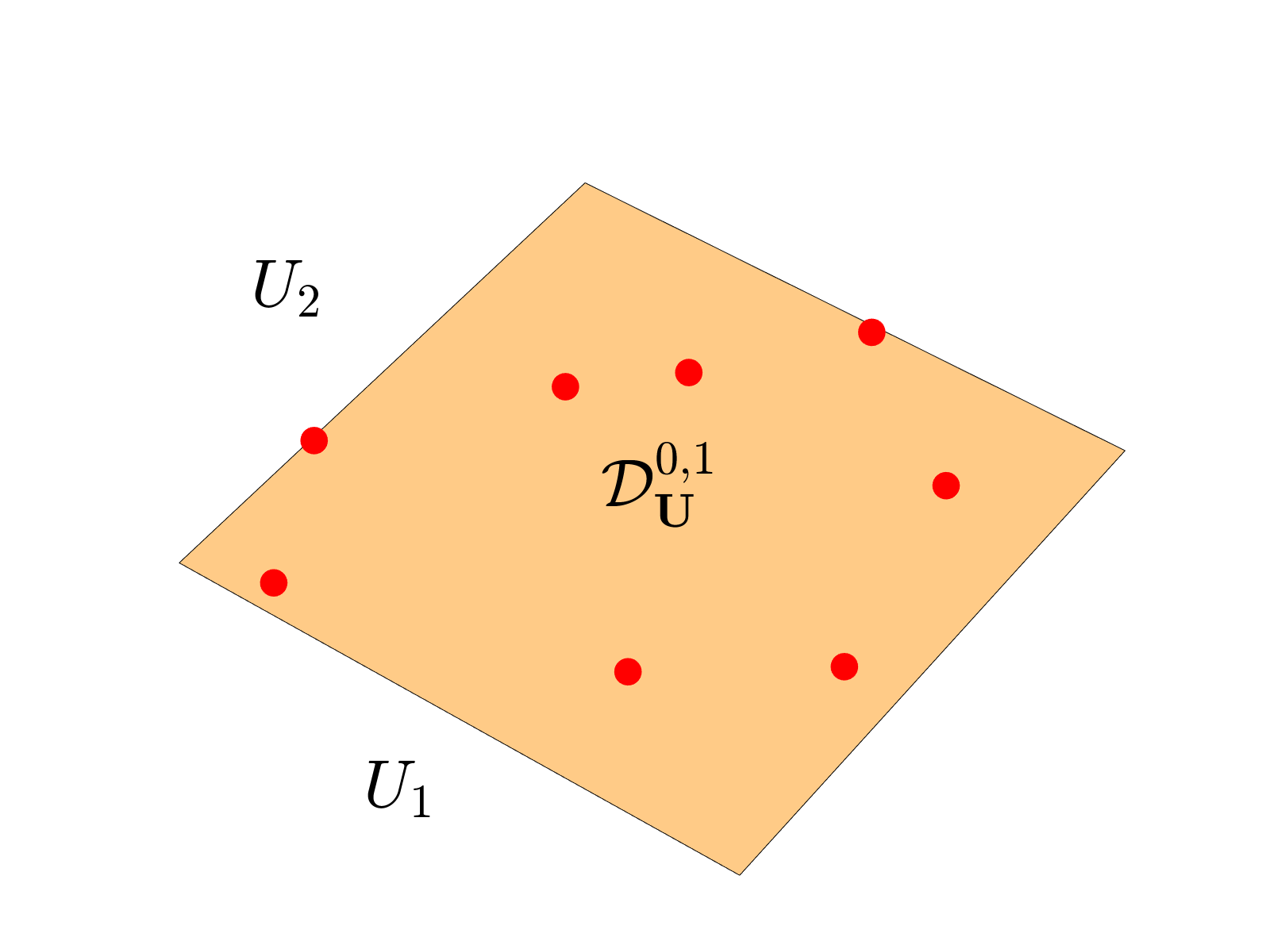}
			
			\includegraphics[width=\linewidth,clip=true,trim=0 0 0 0]{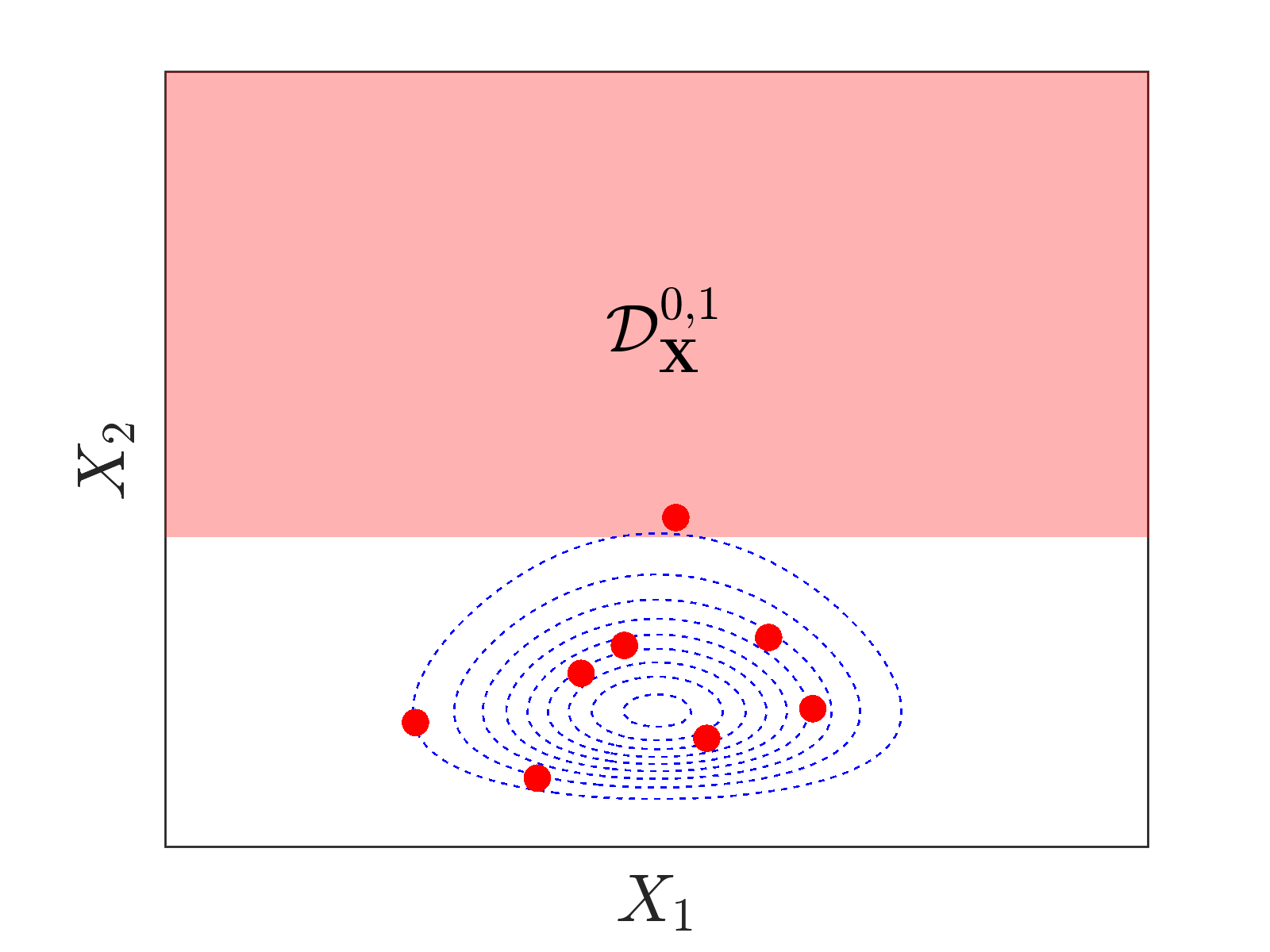}
		\end{minipage}
		\label{fig:SSE partitioning:a}
	}%
	\subfloat[First iteration]{
		\begin{minipage}{0.33\linewidth}
			\includegraphics[width=\linewidth,clip=true,trim=0 0 0 0]{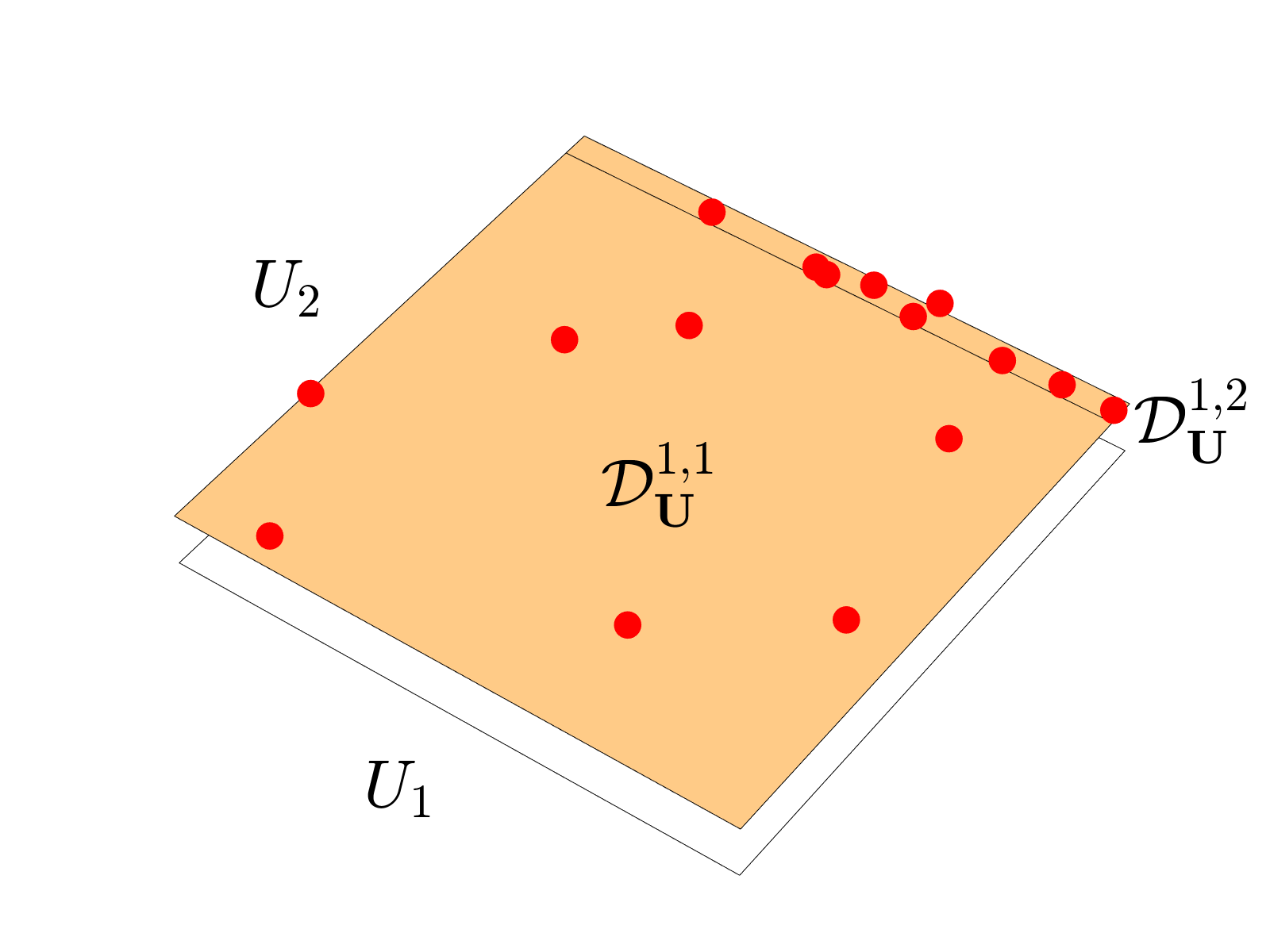}
			
			\includegraphics[width=\linewidth,clip=true,trim=0 0 0 0]{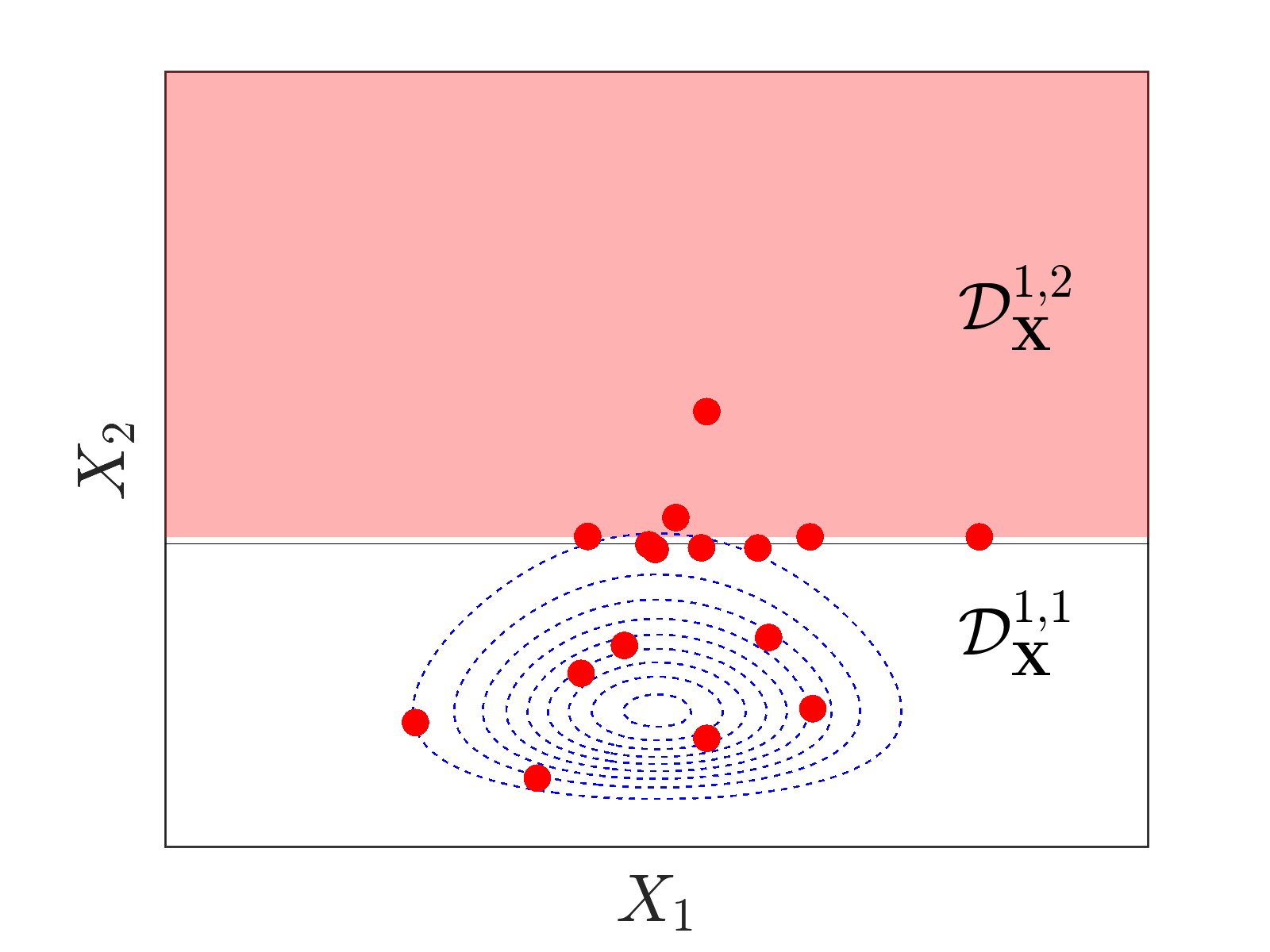}
		\end{minipage}
		\label{fig:SSE partitioning:b}
	}
	\subfloat[Second iteration]{
		\begin{minipage}{0.33\linewidth}
			\includegraphics[width=\linewidth,clip=true,trim=0 0 0 0]{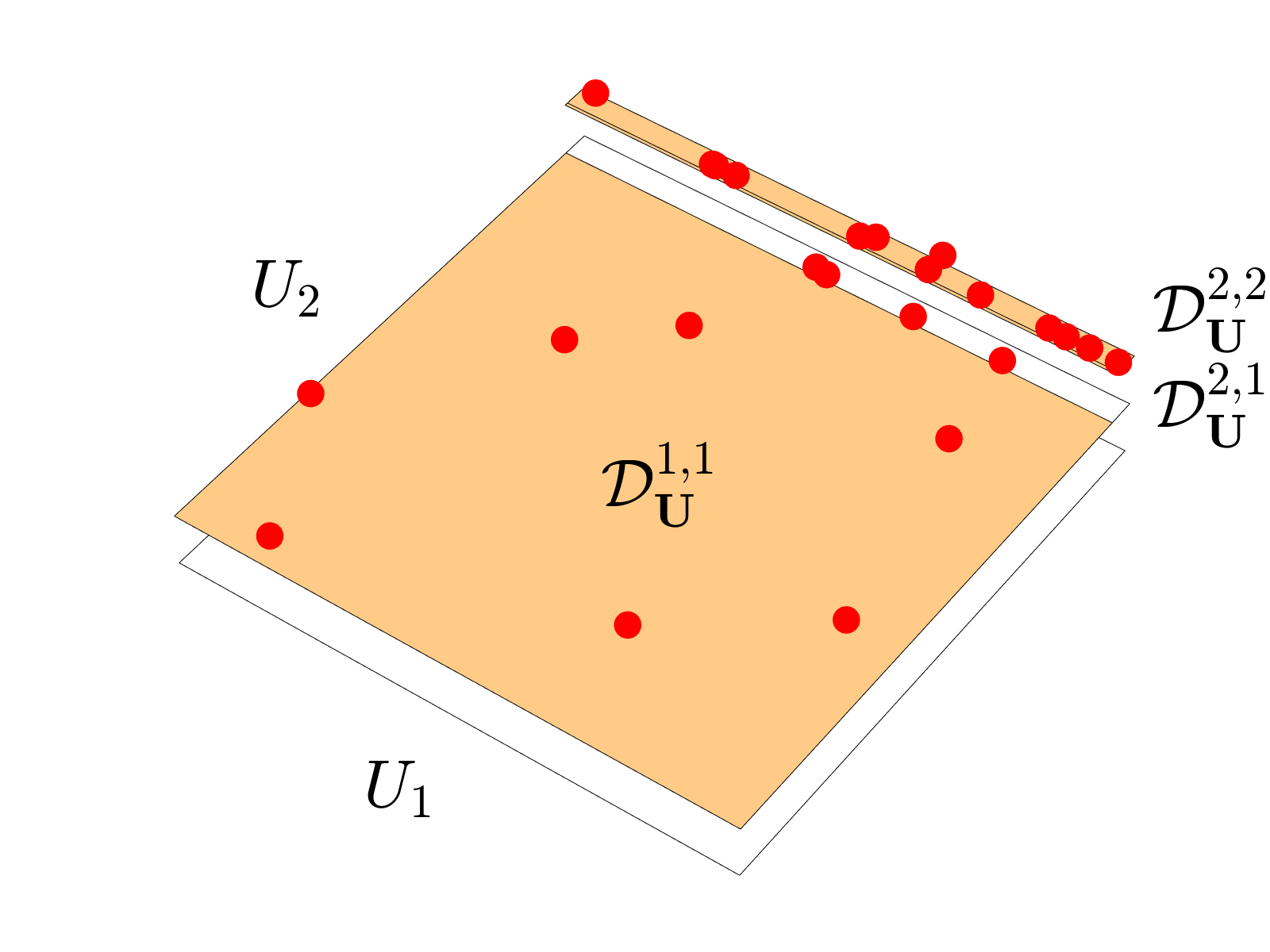}
			
			\includegraphics[width=\linewidth,clip=true,trim=0 0 0 0]{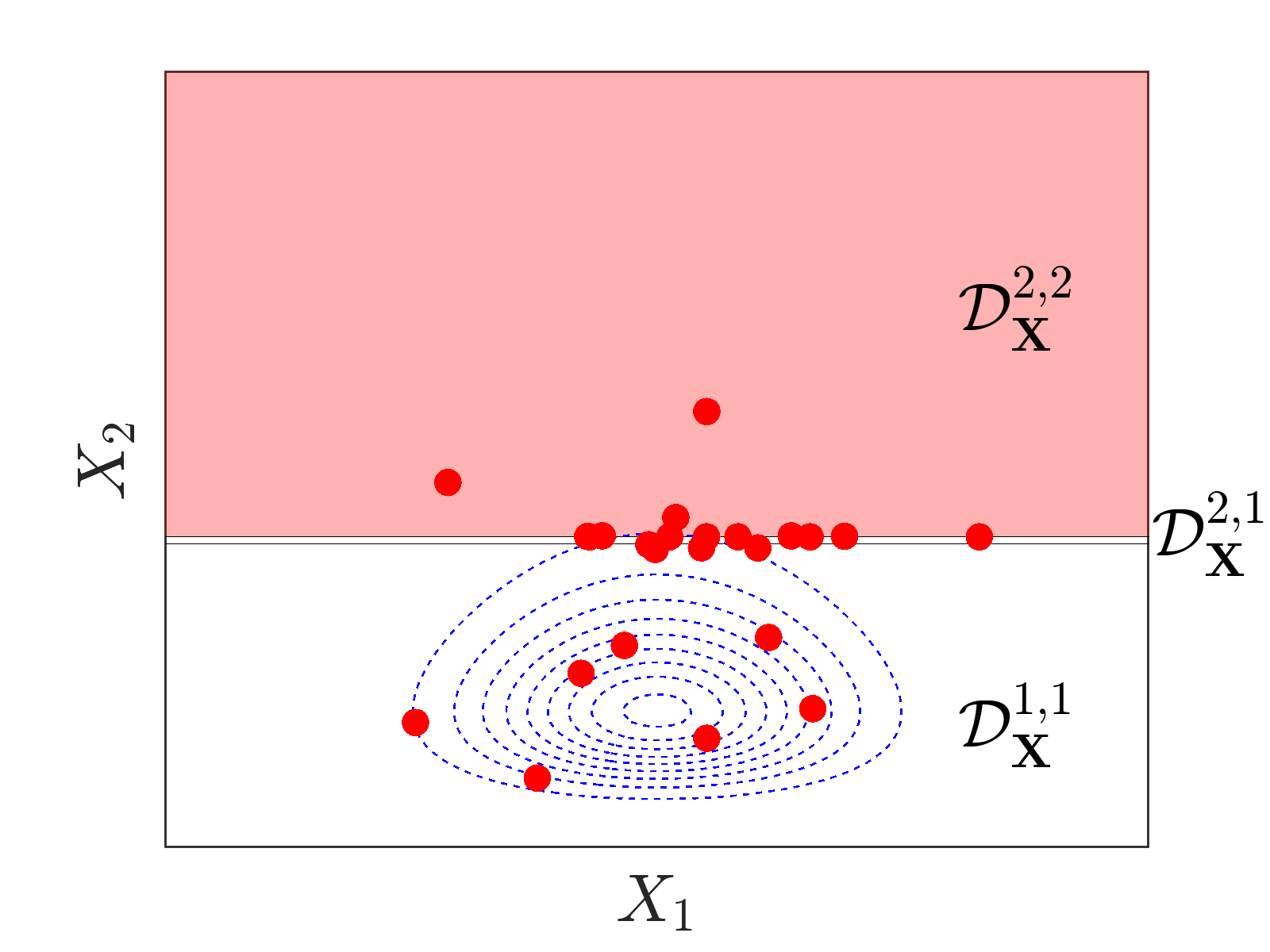}
		\end{minipage}
		\label{fig:SSE partitioning:c}
	}%
	\caption{Graphical representation of the first steps of the active learning algorithm described in Section~\ref{sec:changesAlgo:algorithm} for a two-dimensional toy problem with independent inputs. Upper row: partitioning in the quantile space; lower row: partitioning in the unbounded real space with $f_{\vX}$ contour lines in dashed blue. Red dots show the adaptive experimental design with $\NRefine=4$. The terminal domains $\cT$ are highlighted in orange. The red area denotes the failure domain. In this illustrative example the failure probability is $\hat{P}_f \approx 4.2\cdot 10^{-2}$.}
	\label{fig:SSE partitioning}
\end{figure}

A graphical sketch of the algorithm for a two-dimensional toy problem is shown in Figure~\ref{fig:SSE partitioning}. \rev{The figure shows the first two iterations of the SSE construction in the quantile and real space, highlighting the terminal domains $\cT$ at each step in orange. The initialization step in Figure~\subref{fig:SSE partitioning:a} shows the initial experimental design and global domain $\cD_{\vX}^{0,1}$ and their counterpart in the quantile space $\cD_{\vU}^{0,1}$. At the first iteration, the algorithm chooses the only available refinement domain $k_{\mathrm{refine}}=(0,1)$, partitions it into two subdomains, enriches the experimental design and constructs two additional residual expansions. At the second iteration in Figure~\subref{fig:SSE partitioning:c}, the algorithm refines $\cD_{\vX}^{(1,2)}$ out of two terminal domains $\cT=\{(1,1),(1,2)\}$, because it has the largest contribution to the total failure probability variance. Refinement then continues until the stopping criterion is met.}

\subsubsection{Stopping criterion}
\label{sec:changesAlgo:convergence}
The algorithm can be terminated based on any stopping criterion from the active learning literature \citep{Moustapha2021}. In the present implementation, we opt for a criterion based on the stability of the reliability index bounds, also known as \emph{beta bounds}. The generalized reliability index can be computed from the failure probability as $\beta=-\Phi^{-1}(P_f)$, where $\Phi^{-1}$ is the inverse normal CDF \cite{Ditlevsen1979}. Using the $95\%$ confidence intervalls on $\hat{\beta}$ from the bootstrap replications as defined for the failure probability in Eq.~\eqref{eq:failureBounds}, the stopping criterion is given by
\begin{equation}
	\label{eq:stoppingCriterion}
	\frac{\overline{\hat{\beta}}-\underline{\hat{\beta}}}{\hat{\beta}} < \varepsilon_{\hat{\beta}},
\end{equation}
where the failure threshold is heuristically set to $\varepsilon_{\hat{\beta}}=3\%$. For stability, we abort the algorithm only if this criterion is fulfilled in three consecutive iterations.

Additionally, the algorithm is terminated if the computational budget $\Ntot$, corresponding to the total number of admissible limit-state evaluations, has been exhausted.

\subsection{\rev{Conditional} failure probability estimation}
\label{sec:estimatingLocFail}

In the SSE approach, the total failure probability is decomposed into a sum of weighted, \rev{conditional} failure probabilities (see Eq.~\eqref{eq:failureProb_SSE}). As a side effect of the proposed partitioning strategy (Section~\ref{sec:changesAlgo:partition}), these \rev{conditional} failure probabilities typically end up being quite large or negligibly small. 

Because of the SSE surrogate, simulation methods can be used at a low computational cost. The case of large \rev{conditional} failure probabilities can be treated easily with simple Monte Carlo simulation. The case of small \rev{conditional} failure probabilities, however, is more challenging even with surrogate models. Small failure probabilities should not be underestimated by the algorithm in large domains, because they can contribute significantly to the total failure probability. In cases where the Monte Carlo simulation does not detect failure domains with the provided computational budget, we therefore switch to subset simulation \citep{Au2001} that is known to effectively estimate very small failure probabilities \citep{Moustapha2021}.

\section{Applications}
\label{sec:Applications}

In this section, the proposed SSER algorithm is applied to four problems of varying complexity. These applications were selected to benchmark the algorithm performance in different types of reliability problems: the first two applications have analytical limit-state functions that are challenging for many standard reliability methods. They are the well-known \emph{four-branch function} with two input parameters and four distinct failure regions, and a two-dimensional \emph{piecewise linear} reliability problem that is known to be challenging for subset simulation. The third and fourth applications are engineering models that involve finite element simulations. The third application is an engineering model of a \emph{five-story frame} with $M=21$ mutually dependent input parameters. Lastly, the fourth application is a continuum mechanical problem of a \emph{plate with a hole}, where the Young's modulus is parametrized by a high dimensional ($M=869$) random field.

All computations were done in MATLAB with the UQLab uncertainty quantification framework \cite{MarelliUQLab2014}. Specifically, we used the input module \cite{UQdoc_14_102} for modeling the random inputs, the PCE module \cite{UQdoc_14_104} for the domain-wise spectral expansions with bootstrap replications and the reliability module \cite{UQdoc_14_107} for computing the \rev{conditional} failure probabilities.

A summary of the results and comparisons to other state-of-the-art reliability algorithms is given in Table~\ref{tab:Application:Summary}, and a detailed rundown of each is provided in the following.

\begin{table}
	\begin{minipage}{\linewidth}
		\renewcommand*{\thempfootnote}{\alph{mpfootnote}}
		\renewcommand*{\thefootnote}{\alph{footnote}}
		\setcounter{mpfootnote}{0}
		\setcounter{footnote}{0}
		\centering
		\footnotesize
		\addtolength{\tabcolsep}{0pt}
		\renewcommand{\arraystretch}{1.1}
		
		\begin{tabularx}{\linewidth}{llXXXX}
			\hline
			& & \textbf{Four-branch} & \textbf{Piecewise linear} & \textbf{Five-storey frame} & \textbf{Plate with a hole}\\
			& & Section~\ref{sec:Applications:FourBranch} & Section~\ref{sec:Applications:PiecewiseLinear} & Section~\ref{sec:Applications:5StoryFrame} & Section~\ref{sec:Applications:PlateWithHole}\\
			\hline
			\multirow{6}{*}{\begin{sideways}Reference\end{sideways}} & Method & MCS & MCS & MCS & SuS \cite{Uribe2020}\\
			& $\hat{P}_f$ & $4.46\cdot 10^{-3}$ & $3.2\cdot 10^{-5}$ & $1.49\cdot 10^{-6}$ & $3.75\cdot 10^{-6}$\\
			& $[\underline{\hat{P}_f};\overline{\hat{P}_f}]$ & $[4.42;4.51]\cdot 10^{-3}$ & $[3.1;3.3]\cdot 10^{-5}$ & $[1.1;1.8]\cdot 10^{-6}$ & $[3.59;3.91]\cdot 10^{-6}$\footnotemark\\
			& $\hat{\beta}$ & $2.62$ & $4.00$ & $4.67$ & $4.48$\\
			& $[\underline{\hat{\beta}};\overline{\hat{\beta}}]$ & $[2.61;2.62]$ & $[3.99;4.00]$ & $[4.64;4.73]$ & $[4.47;4.48]$\\
			& $N$ & $10^7$ & $10^8$ & $10^8$ & $1.62\cdot 10^{6}$\\
			\hline
			\multirow{6}{*}{\begin{sideways}Competing algorithms\end{sideways}}& Method & AK-MCS+U \cite{Echard2011} & -- & sPCE+MCS \cite{Blatman2010Adaptive}\footnotemark
			& iCEred \cite{Uribe2020}\\
			& $\hat{P}_f$ & $4.42\cdot 10^{-3}$ & -- & $3.9\cdot 10^{-7}$ & $3.57\cdot 10^{-6}$\\
			& $[\underline{\hat{P}_f};\overline{\hat{P}_f}]$ & -- & -- & -- & $[3.21;3.93]\cdot 10^{-6}$\\
			& $\hat{\beta}$ & $2.62$ & -- & $4.94$ & $4.49$\\
			& $[\underline{\hat{\beta}};\overline{\hat{\beta}}]$ & -- & -- & -- & $[4.47;4.51]$\\
			& $N$ & $126$ & -- & $450$ & $2{,}396$\cite{Uribe2020}\footnotemark
			\\
			\hline
			\multirow{7}{*}{\begin{sideways}Proposed algorithm\end{sideways}}& Method & SSER & SSER & SSER & SSER\\
			& $\hat{P}_f$ & $4.44\cdot 10^{-3}$ & $3.16\cdot 10^{-5}$ & $1.49\cdot 10^{-6}$ & $3.27\cdot 10^{-6}$\\
			& $[\underline{\hat{P}_f};\overline{\hat{P}_f}]$ & $[4.43;4.47]\cdot 10^{-3}$ & $[3.16;3.16]\cdot 10^{-5}$ & $[1.20;1.64]\cdot 10^{-6}$ & $[3.26;5.47]\cdot 10^{-6}$\\
			& $\hat{\beta}$ & $2.62$ & $4$ & $4.67$ & $4.50$\\
			& $[\underline{\hat{\beta}};\overline{\hat{\beta}}]$ & $[2.60;2.63]$ & $[4;4]$ & $[4.65;4.71]$ & $[4.40;4.51]$\\
			& $N$ & $270$ & $480$ & $480$ & $ 1{,}400$\\
			& $[\underline{N};\overline{N}]$\footnotemark
			& $[180;330]$ & $[320;720]$ & $[320;720]$ & $[1{,}200;2{,}000]$ \\
			\hline		
		\end{tabularx}
		\caption{\emph{Applications summary}: if not given with an external reference, the values were computed with UQLab \cite{MarelliUQLab2014,UQdoc_14_107}. Values in square brackets correspond to the $95\%$ confidence intervals. $N$ denotes the number of limit-state evaluations needed to compute the reported values. For SSER confidence intervals are also given for the number of limit-state evaluations until convergence from a set of independent runs. The used methods are Monte Carlo simulation (MCS), subset simulation (SuS), adaptive kriging Monte Carlo simulation with $U$ learning function (AK-MCS+U), sparse polynomial chaos expansions with Monte Carlo simulation (sPCE+MCS), improved CE method with failure-informed dimension reduction (iCEred). \label{tab:Application:Summary}}
		\footnotetext[1]{Estimated from $100$ independent SuS runs with reported $\mathrm{CV}_{\hat{P}_f}=0.215$ as $\hat{P}_f\cdot(1\pm2\mathrm{CV}_{\hat{P}_f}/\sqrt{100})$.}
		\footnotetext[2]{Reported value $\hat{P}_f=3.9\cdot 10^{-7}$ does not lie within reference confidence intervals.}
		\footnotetext[3]{Additional $675$ gradient evaluations.}
		\footnotetext[4]{From independent algorithm runs to reach convergence.}
	\end{minipage}
\end{table}

\subsection{Four-branch function}
\label{sec:Applications:FourBranch}

The four-branch function is a common benchmark in the active learning literature \cite{Schueremans2005, Echard2011}. It simulates a series system with four distinct limit-state surfaces. The input random parameter is distributed according to a bivariate standard normal distribution, \ie $\vX\sim\mathcal{N}(\ve{0},\ve{1}_{2})$. The limit-state function is given analytically as
\begin{equation}
	\label{ex1:ls}
	g(\vX) = \min
	\begin{cases}
		3 + 0.1(X_1-X_2)^2 - \frac{X_1+X_2}{\sqrt{2}}\\
		3 + 0.1(X_1-X_2)^2 + \frac{X_1+X_2}{\sqrt{2}}\\
		X_1-X_2 + \frac{6}{\sqrt{2}}\\
		X_2-X_1 + \frac{6}{\sqrt{2}}.\\
	\end{cases}
\end{equation}

The reference failure probability of this problem from Monte Carlo simulation is $P_f=4.46\cdot 10^{-3}$ ($\beta=2.62$) obtained with $N=10^7$ limit-state evaluations.

We perform $50$ independent runs of SSER on this problem, with different random seeds and initial designs. 
The number of refinement samples was set to $\NRefine=15$ and the maximum polynomial degree for the residual expansions to $p_{\mathrm{max}}=2$. The domain-wise accuracy was estimated with $B=500$ bootstrap replications. 

\begin{figure}
	\centering
	\subfloat[Single run]{
		\begin{minipage}{0.49\linewidth}
			\includegraphics[width=\linewidth,clip=true,trim=0 0 0 0]{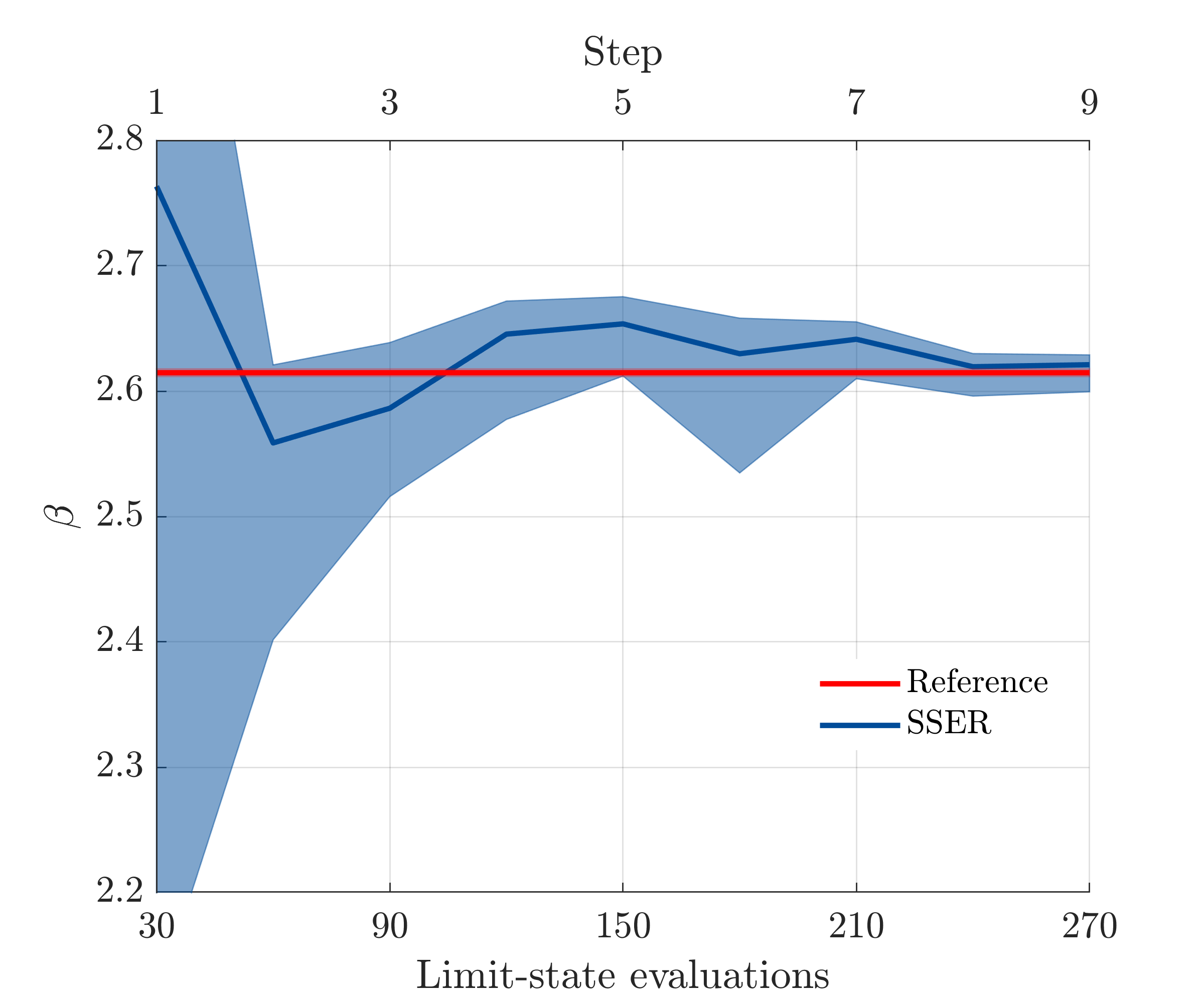}
		\end{minipage}
		\label{fig:Ex1:convergence:single}
	}%
	\subfloat[$50$ independent runs]{
		\begin{minipage}{0.49\linewidth}
			\includegraphics[width=\linewidth,clip=true,trim=0 0 0 0]{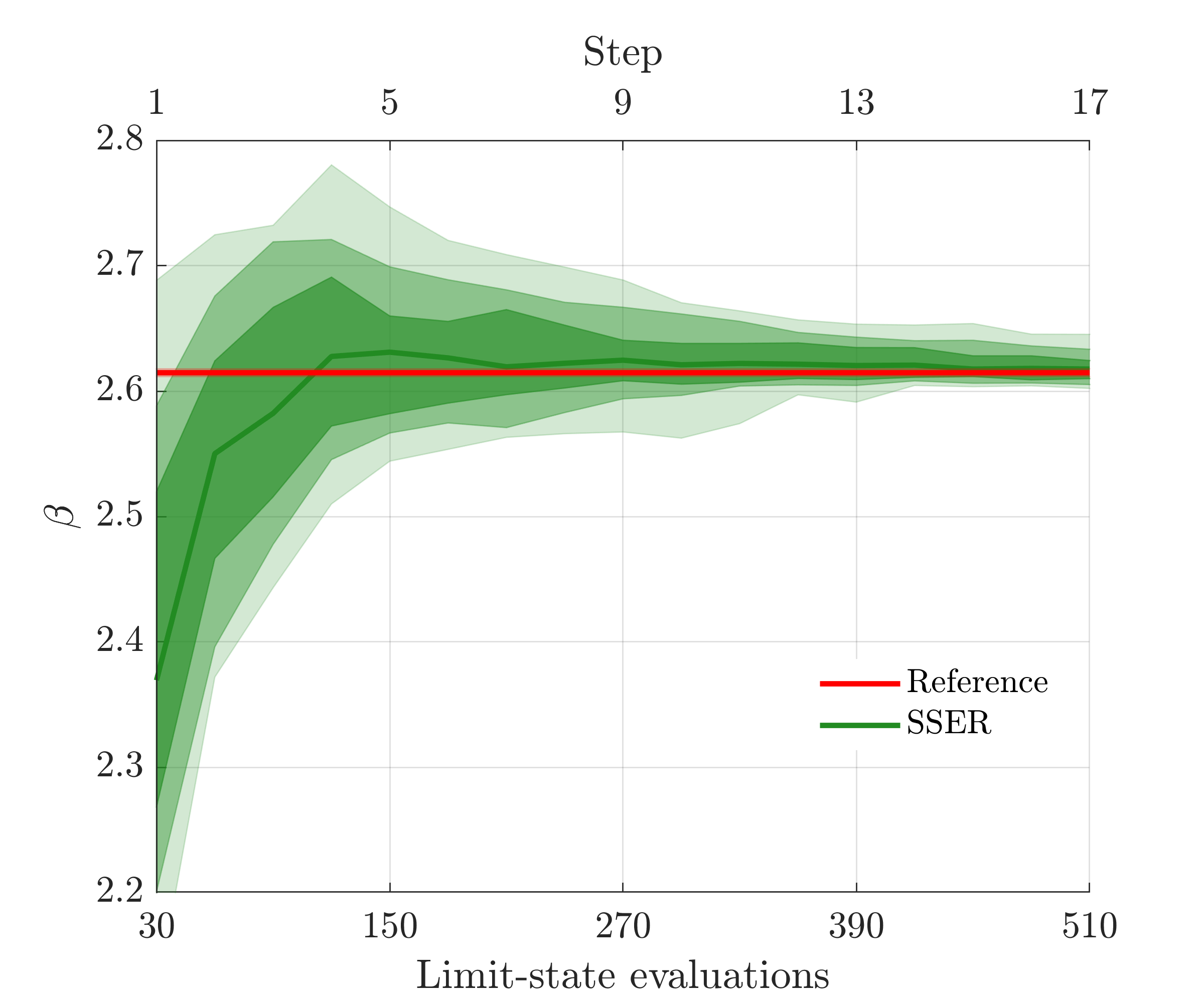}
		\end{minipage}
		\label{fig:Ex1:convergence:replications}
	}
	\caption{\emph{Four-branch function}: Convergence to the reference reliability index $\beta$ as a function of limit-state evaluations and algorithm steps. The $95\%$ confidence intervals in Figure~\protect\subref{fig:Ex1:convergence:single} are based on the bootstrap replications. This figure shows the convergence until the stopping criterion in Eq.~\eqref{eq:stoppingCriterion} is met at Step~$9$ of SSER. The bounds in Figure~\protect\subref{fig:Ex1:convergence:replications} are the $90\%$, $75\%$ and $50\%$ confidence intervals from the mean predictions of the independent SSER runs. \label{fig:Ex1:convergence}}	
\end{figure}

The convergence to the reference reliability index for a selected run of SSER and all $50$ independent runs is shown in Figure~\ref{fig:Ex1:convergence}. The median number of steps required to reach the stopping criterion is $7$, corresponding to $\underbrace{2\NRefine}_{\text{initial}} + (6\cdot 2) \NRefine=210$ limit-state evaluations.

\begin{figure}
	\centering
	\subfloat[Step $1$]{
		\begin{minipage}{0.49\linewidth}
			\includegraphics[width=\linewidth,clip=true,trim=0 0 0 0]{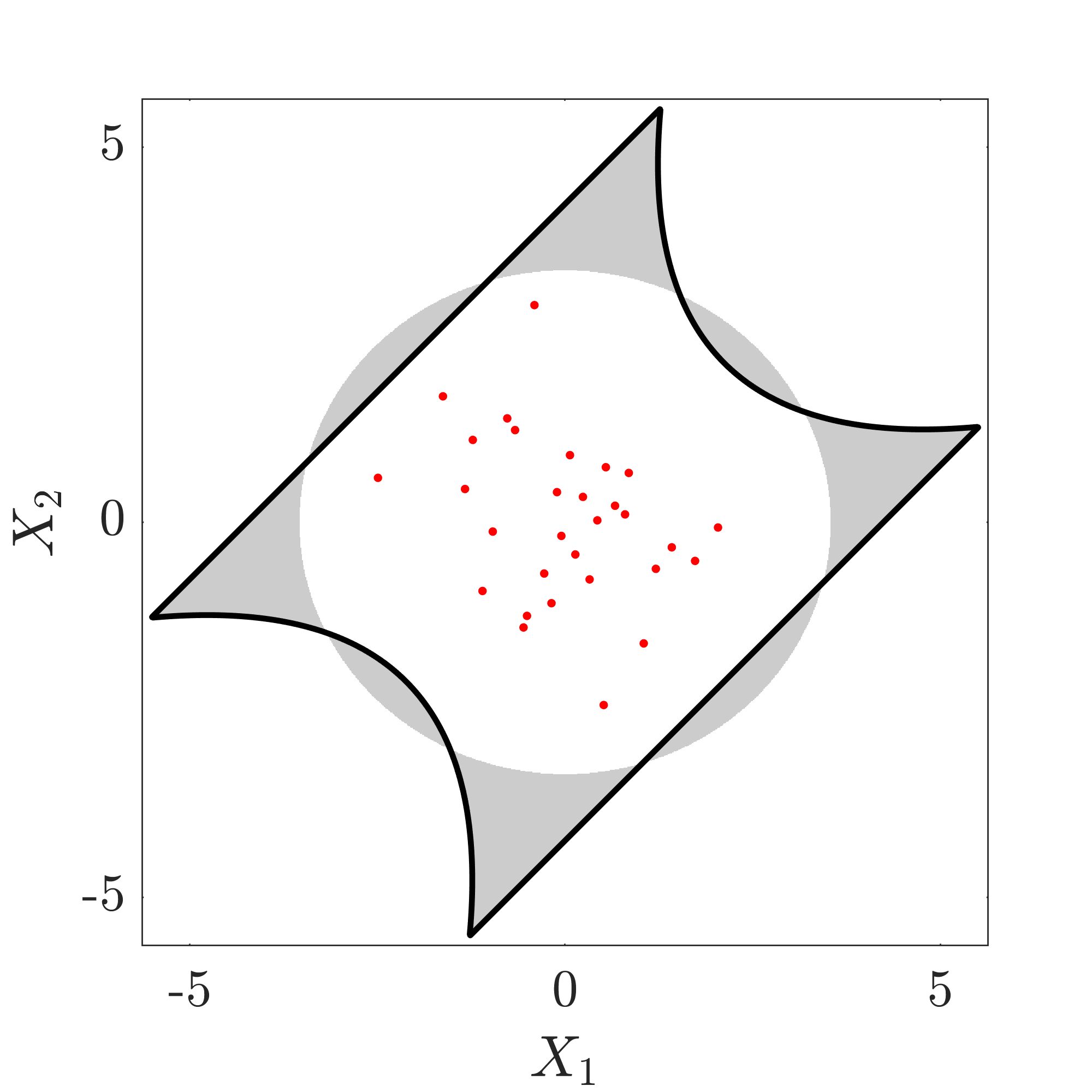}
		\end{minipage}
		\label{fig:Ex1:error:1}
	}%
	\subfloat[Step $2$]{
		\begin{minipage}{0.49\linewidth}
			\includegraphics[width=\linewidth,clip=true,trim=0 0 0 0]{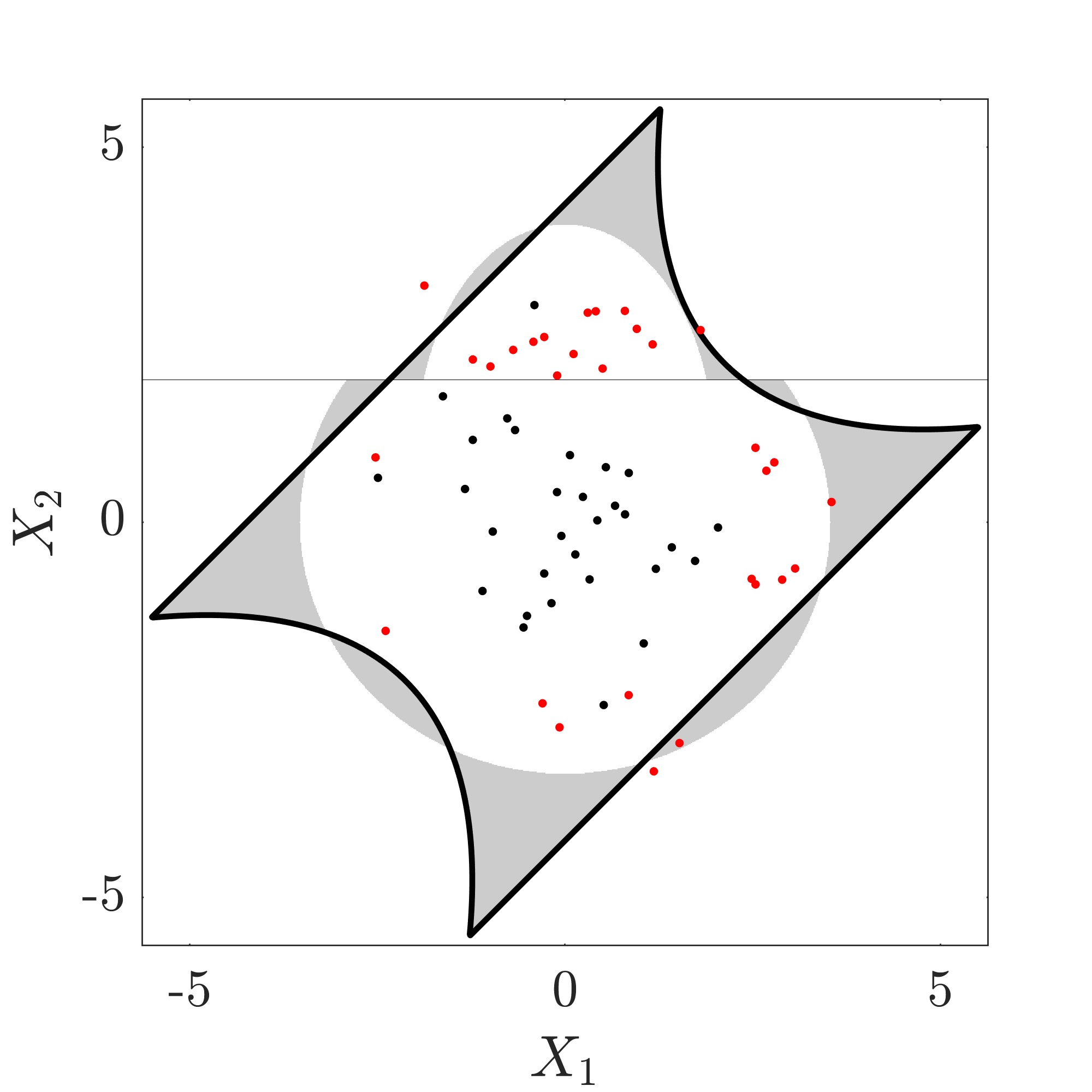}
		\end{minipage}
		\label{fig:Ex1:error:2}
	}
	
	\subfloat[Step $6$]{
		\begin{minipage}{0.49\linewidth}
			\includegraphics[width=\linewidth,clip=true,trim=0 0 0 0]{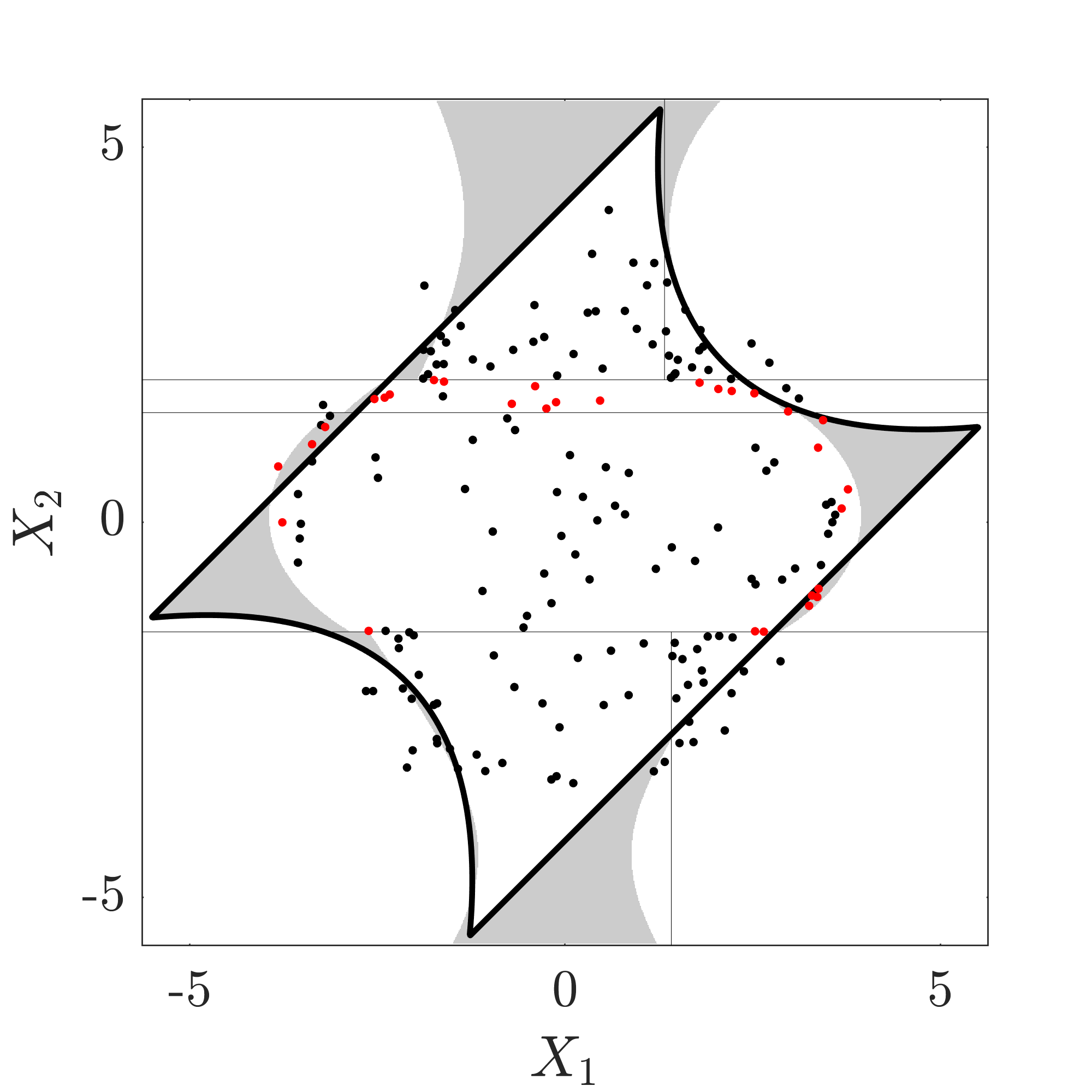}
		\end{minipage}
		\label{fig:Ex1:error:3}
	}%
	\subfloat[Step $9$]{
		\begin{minipage}{0.49\linewidth}
			\includegraphics[width=\linewidth,clip=true,trim=0 0 0 0]{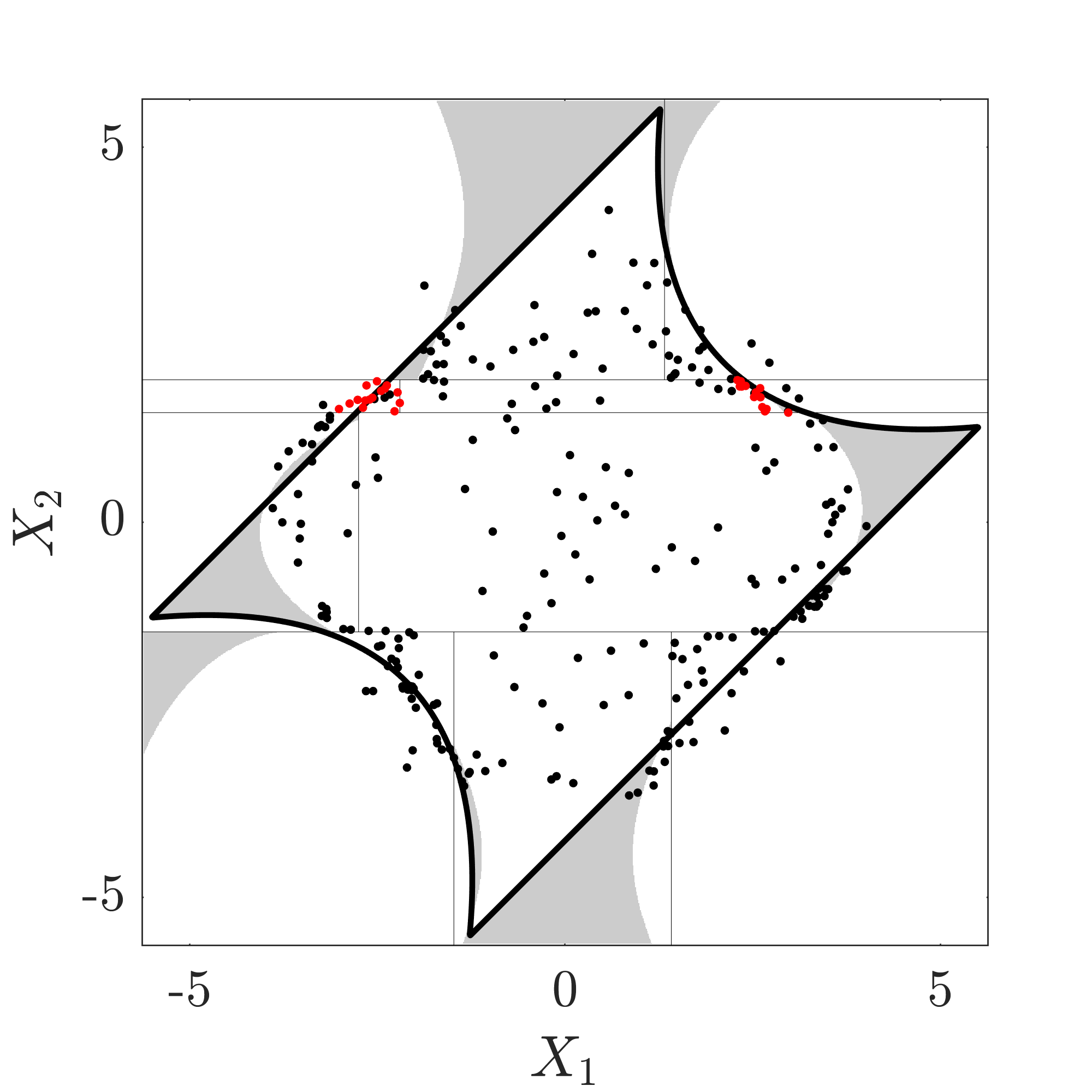}
		\end{minipage}
		\label{fig:Ex1:error:4}
	}
	\caption{\emph{Four-branch function}: Misclassification of failure/safe points for selected steps of SSER. The white areas were classified correctly, while the gray areas were misclassified. The thick black line indicates the limit-state surface. The domain bounds identified by the SSER are shown as thin black lines. The black points are the used experimental design and the red points are the points added at the shown step of the algorithm. The convergence of the reliability index estimator is shown in Figure~\protect\subref{fig:Ex1:convergence:single}. \label{fig:Ex1:error}}
\end{figure}

A more in-depth look at the SSER performance is shown in Figure~\ref{fig:Ex1:error}, where four selected steps of the algorithm run from Figure~\subref{fig:Ex1:convergence:single} are displayed. It is important to note that due to the Gaussian probability measure, the importance of regions shrinks exponentially with the distance from the domain centre (\ie $(0,0)$). The initial approximation is a paraboloid that underestimates the failure probability by misclassifying regions near the domain centre. At the second step, SSER partitions the initial domain along $d=2$ at $\widehat{X}_2\approx 2$ corresponding to the $98$th percentile (\ie $\widehat{u}_2=0.98$). By Step~$6$ the overall approximation is extremely accurate near the centre of the domain, while it remains poor near the unimportant four corners of the safe domain. Three steps later, at Step~$9$, the stopping criterion is met and the algorithm is terminated.

\subsection{Piecewise linear function}
\label{sec:Applications:PiecewiseLinear}

The \emph{piecewise linear function} was devised in \citet{Breitung2019} to break the subset simulation algorithm \cite{Au2001}. This function has two limit-state surfaces, one with a large and the second with a negligible failure probability. By providing a steep limit-state function in the direction of the second limit-state surface, the subset simulation algorithm overlooks the first limit-state surface. 

The input random parameter is again distributed according to a bivariate standard normal distribution, \ie $\vX\sim\mathcal{N}(\ve{0},\ve{1}_2)$, and the limit-state function is given analytically by
\begin{equation}
	\label{ex2:ls}
	g(\vX) = \min
	\begin{cases}
		g_1(\vX) = 
		\begin{cases}
			4-X_1, &\text{if} \quad X_1>3.5;\\
			0.85-0.1\cdot X_1, &\text{if} \quad X_1\le 3.5;\\
		\end{cases}\\
		g_2(\vX) = 
		\begin{cases}
			0.5-0.1\cdot X_2,&\text{if} \quad X_2>2;\\
			2.3-X_2,&\text{if} \quad X_2\le 2.\\
		\end{cases}
	\end{cases}
\end{equation}

We obtain a reference failure probability of $P_f=3.2\cdot 10^{-5}$ ($\beta=4$) for this problem from Monte Carlo simulation with $N=10^8$ samples.

We perform $50$ independent runs of SSER, with different random seeds. The number of refinement samples was set to $\NRefine=40$ and the maximum polynomial degree for the residual expansions to $p_{\mathrm{max}}=6$. The domain-wise accuracy was estimated with $B=500$ bootstrap replications. 

\begin{figure}
	\centering
	\subfloat[Single run]{
		\begin{minipage}{0.49\linewidth}
			\includegraphics[width=\linewidth,clip=true,trim=0 0 0 0]{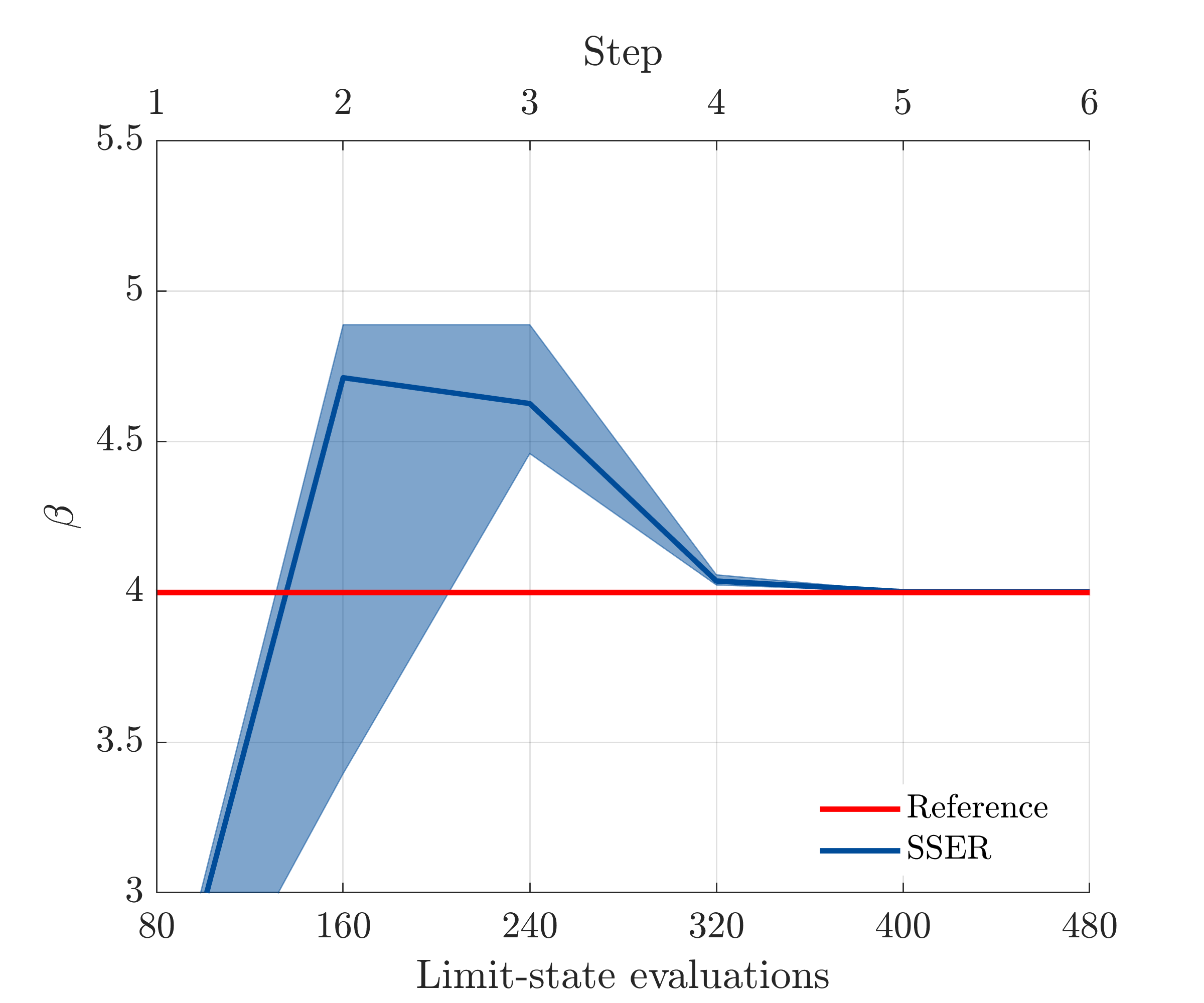}
		\end{minipage}
		\label{fig:Ex2:convergence:single}
	}%
	\subfloat[$50$ independent runs]{
		\begin{minipage}{0.49\linewidth}
			\includegraphics[width=\linewidth,clip=true,trim=0 0 0 0]{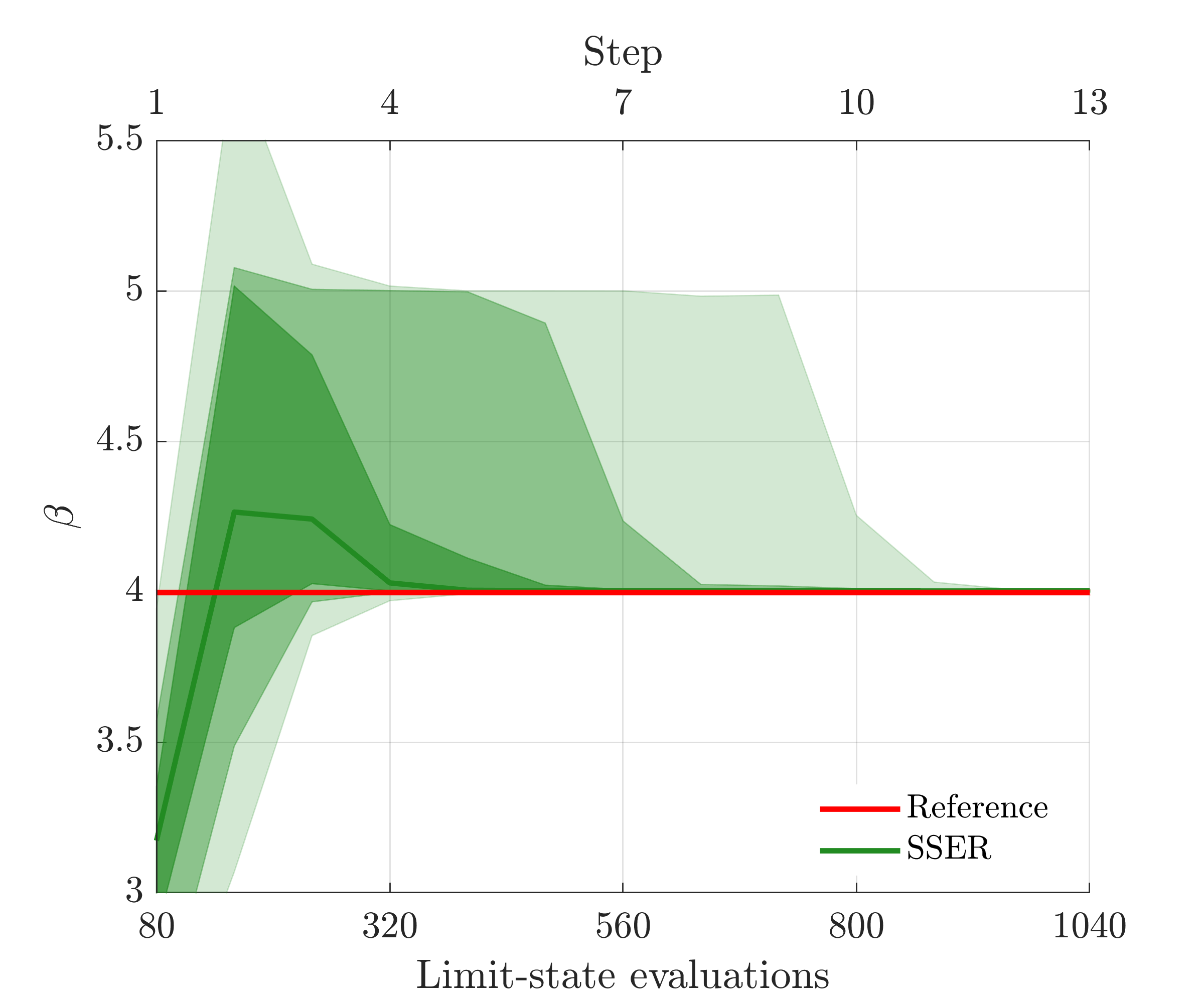}
		\end{minipage}
		\label{fig:Ex2:convergence:replications}
	}
	\caption{\emph{Piecewise linear function}: Convergence to the reference reliability index $\beta$ as a function of limit-state evaluations and algorithm steps. The $95\%$ confidence intervals in Figure~\protect\subref{fig:Ex2:convergence:single} are based on the bootstrap replications. This figure shows the convergence until the stopping criterion in Eq.~\eqref{eq:stoppingCriterion} is met at Step~$6$ of SSER. The bounds in Figure~\protect\subref{fig:Ex2:convergence:replications} are the $90\%$, $75\%$ and $50\%$ confidence intervals from the mean predictions of the independent SSER runs.
		\label{fig:Ex2:convergence}}	
\end{figure}

The convergence of SSER to the reference reliability index is shown in Figure~\ref{fig:Ex2:convergence}. As intended by this limit-state function, most runs of the algorithm initially overestimate the reliability index. Due to its refinement domain selection strategy, however, SSER later returns to the high failure probability domains and accurately estimates the reliability index. The median number of steps required to reach the stopping criterion is $5$, corresponding to $\underbrace{2\NRefine}_{\text{initial}} + (4\cdot 2) \NRefine=400$ limit-state evaluations.

\begin{figure}
	\centering
	\subfloat[Step $1$]{
		\begin{minipage}{0.49\linewidth}
			\includegraphics[width=\linewidth,clip=true,trim=0 0 0 0]{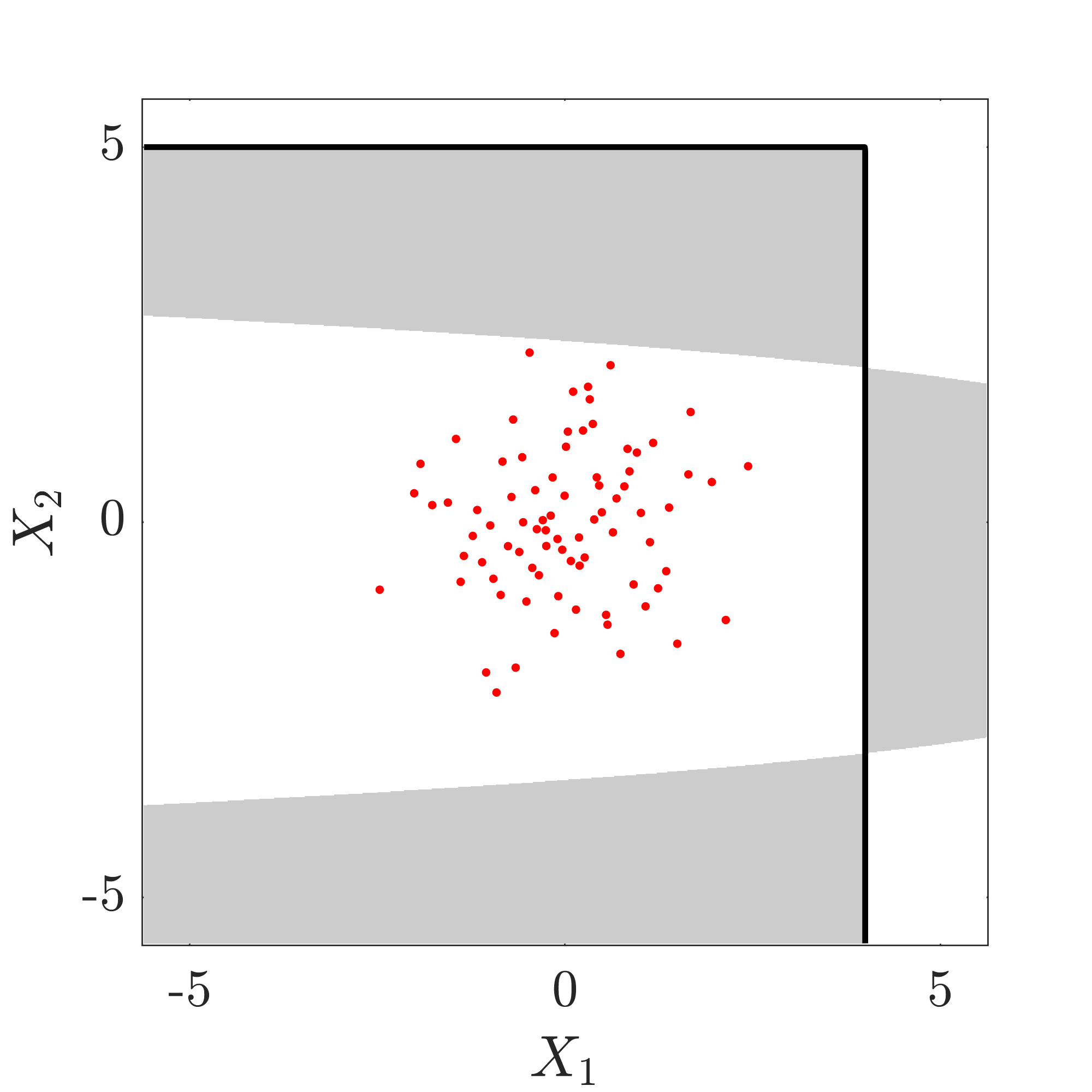}
		\end{minipage}
		\label{fig:Ex2:error:1}
	}%
	\subfloat[Step $2$]{
		\begin{minipage}{0.49\linewidth}
			\includegraphics[width=\linewidth,clip=true,trim=0 0 0 0]{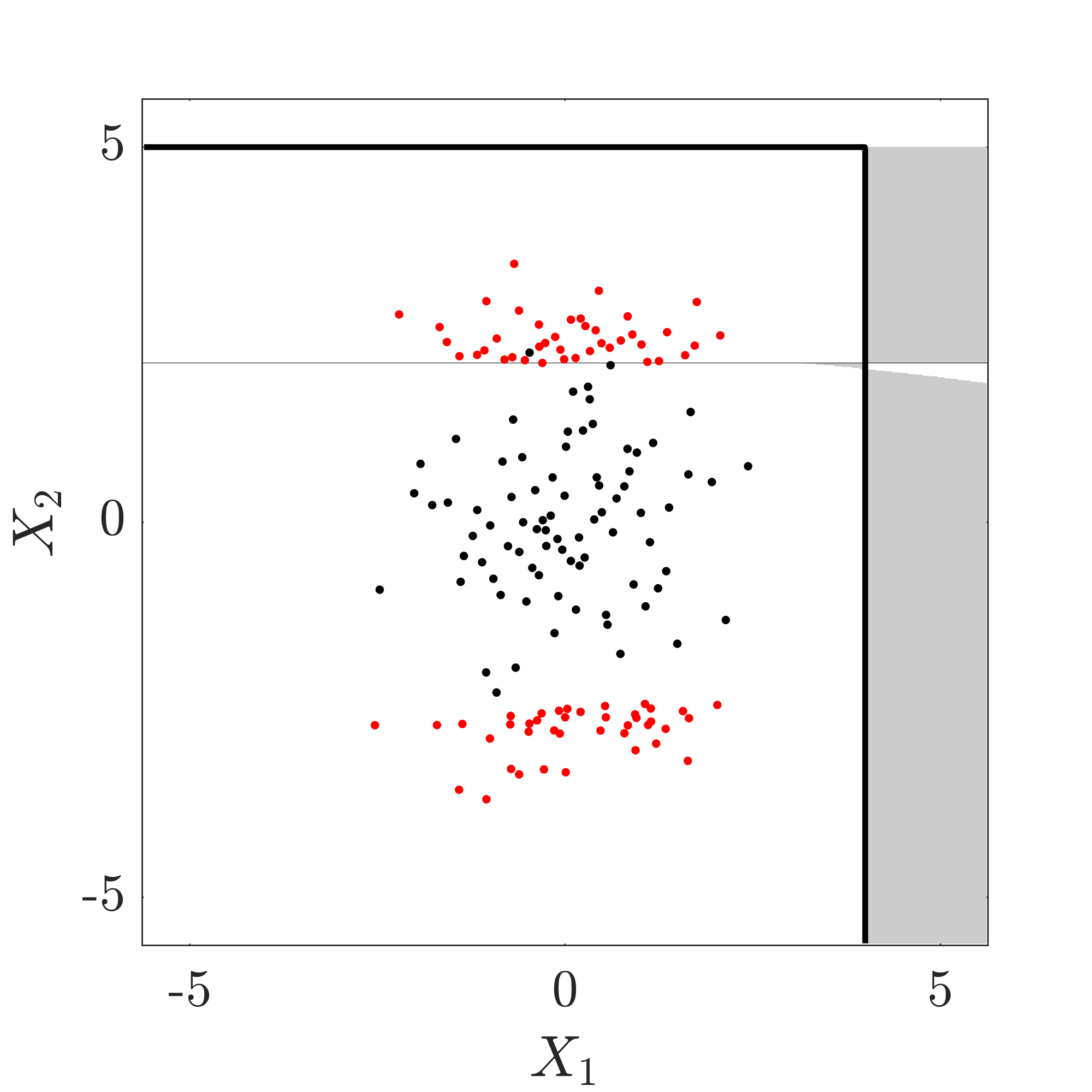}
		\end{minipage}
		\label{fig:Ex2:error:2}
	}
	
	\subfloat[Step $4$]{
		\begin{minipage}{0.49\linewidth}
			\includegraphics[width=\linewidth,clip=true,trim=0 0 0 0]{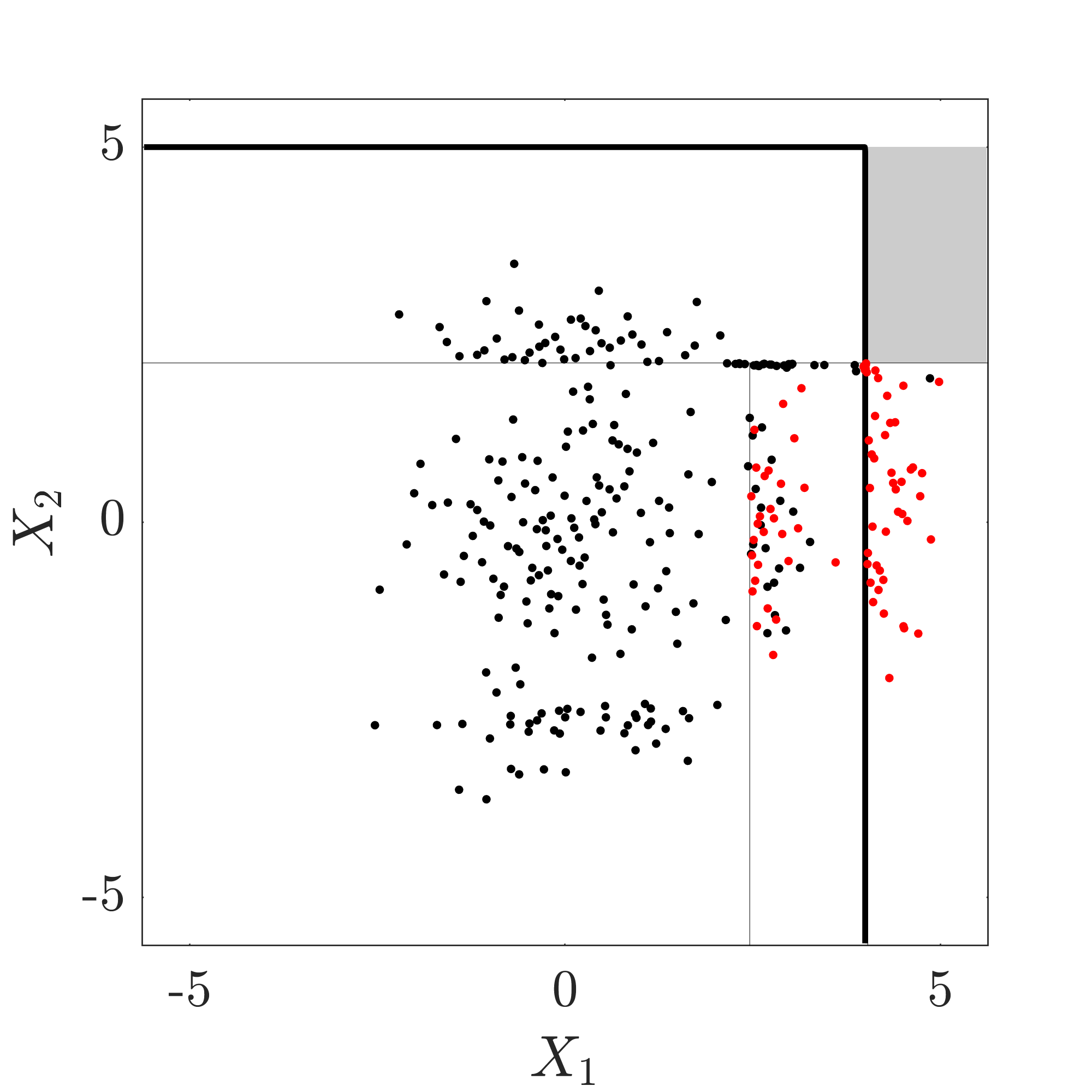}
		\end{minipage}
		\label{fig:Ex2:error:3}
	}%
	\subfloat[Step $6$]{
		\begin{minipage}{0.49\linewidth}
			\includegraphics[width=\linewidth,clip=true,trim=0 0 0 0]{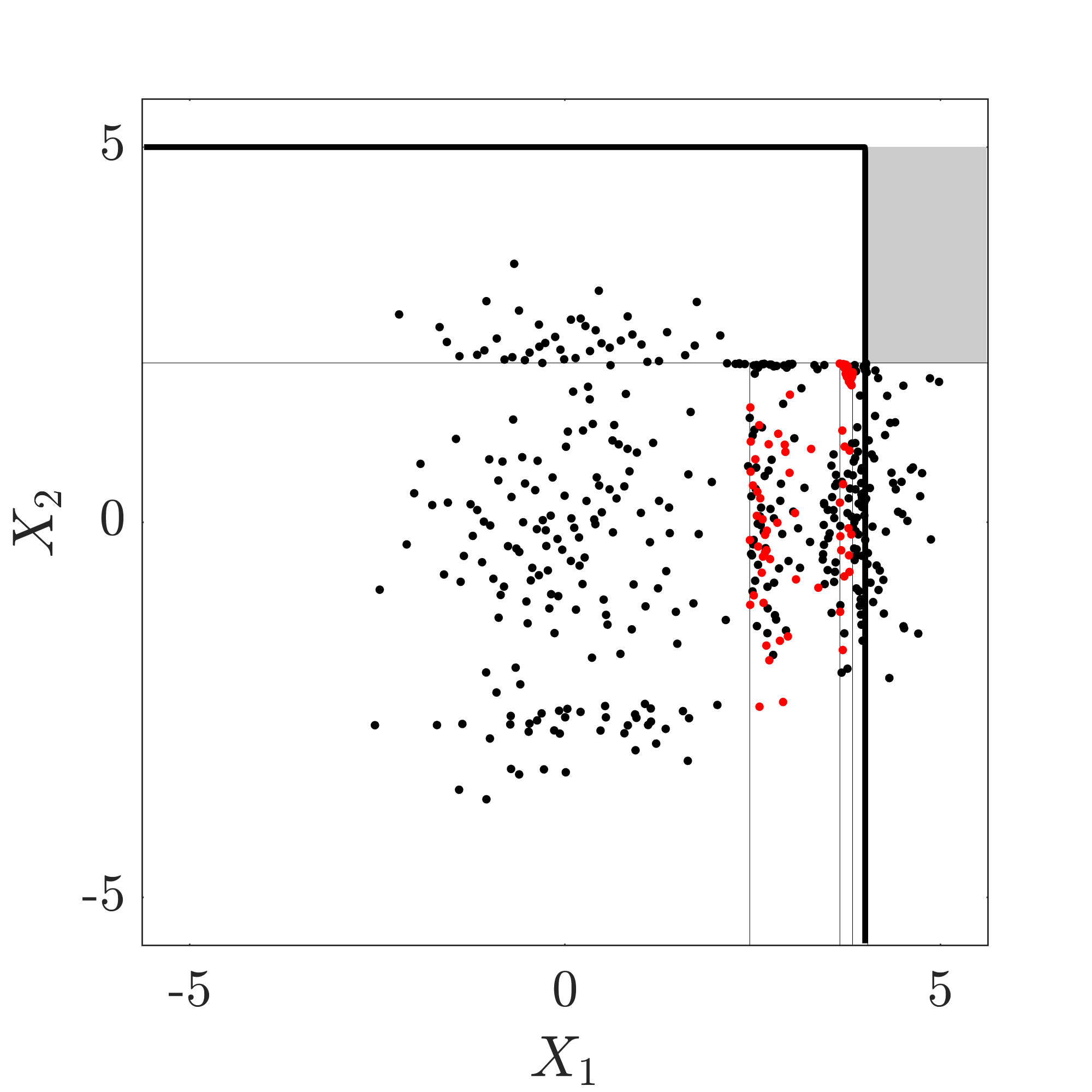}
		\end{minipage}
		\label{fig:Ex2:error:4}
	}
	\caption{\emph{Piecewise linear function}: Misclassification of failure/safe points for selected steps of SSER. The white areas were classified correctly, while the gray areas were misclassified. The thick black line indicates the limit-state surface. The domain bounds identified by SSER are shown as thin black lines. The black points are the used experimental design and the red points are the points added at the shown step of the algorithm. The convergence of the reliability index estimator is shown in Figure~\protect\subref{fig:Ex1:convergence:single}. \label{fig:Ex2:error}}
\end{figure}

Figure~\ref{fig:Ex2:error} reveals why SSER initially overestimates the reliability index and underestimates the failure probability. After the initial approximation, at Step~$2$, the algorithm partitions along $d=2$ to isolate the region where it suspects the highest misclassification probability. In the following steps, it focuses again on the big region that incorporates the larger failure domain and accurately approximates it at Step~$4$. Two steps later, at Step~$6$, the stopping criterion is fulfilled and SSER is terminated.

\subsection{Five-story frame}
\label{sec:Applications:5StoryFrame}

\begin{figure}
	\centering
	\begin{tabular}{cc}
		\adjustbox{valign=b}{\begin{tabular}{@{}c@{}}
				\subfloat[Structure]{
					\begin{minipage}{0.49\linewidth}
						\includegraphics[width=\linewidth,clip=true,trim=0 0 0 0]{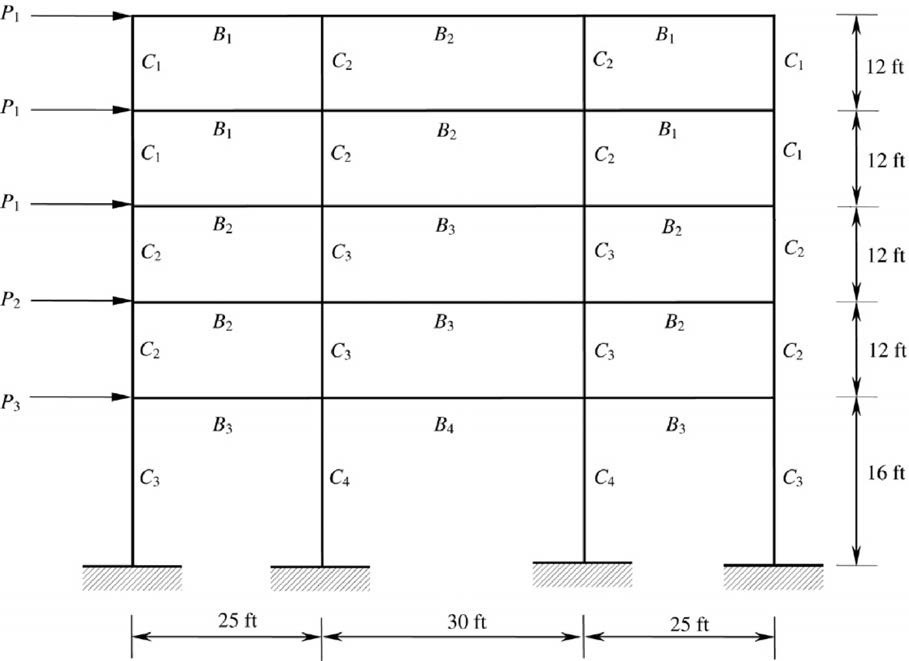}
					\end{minipage}
					\label{fig:Ex3:setup:sketch}
				}\\
				\subfloat[Elements properties]{
					\begin{minipage}{0.49\linewidth}
						\centering
						\footnotesize
						\addtolength{\tabcolsep}{-2pt}
						\renewcommand{\arraystretch}{0.6} 
						\begin{tabular}{cccccccc}
							\hline$B_1$&$B_2$&$B_3$&$B_4$&$C_1$&$C_2$&$C_3$&$C_4$\\
							\hline$E_4$&$E_4$&$E_4$&$E_4$&$E_5$&$E_5$&$E_5$&$E_5$\\
							$I_{10}$&$I_{11}$&$I_{12}$&$I_{13}$&$I_6$&$I_7$&$I_8$&$I_9$\\
							$A_{18}$&$A_{19}$&$A_{20}$&$A_{21}$&$A_{14}$&$A_{15}$&$A_{16}$&$A_{17}$\\
							\hline
						\end{tabular}
					\end{minipage}
					\label{fig:Ex3:setup:elements}
				}
		\end{tabular}}
		&
		\adjustbox{valign=b}{\subfloat[Parameter marginals\label{fig:Ex3:setup:parameters}]{
				\begin{minipage}{0.49\linewidth}
					\footnotesize
					\addtolength{\tabcolsep}{-5pt}
					\renewcommand{\arraystretch}{0.6} 
					\begin{tabular}{cccc}
						\hline\textbf{Parameter}&\textbf{Distribution}\footnote{Gaussians truncated to $[0, +\infty]$}&$\mu$&$\sigma$\\
						\hline$P_1$~(kN)&Lognormal&$133$&$40$\\
						$P_2$~(kN)&Lognormal&$89$&$35.6$\\
						$P_3$~(kN)&Lognormal&$71.2$&$28.5$\\
						$E_4$~(kN/m$^2$)&Gaussian&$2.17\cdot 10^{7}$&$1.92\cdot 10^{6}$\\
						$E_5$~(kN/m$^2$)&Gaussian&$2.38\cdot 10^{7}$&$1.92\cdot 10^{6}$\\
						$I_6$~(m$^4$)&Gaussian&$8.13\cdot 10^{-3}$&$1.08\cdot 10^{-3}$\\
						$I_7$~(m$^4$)&Gaussian&$0.0115$&$1.3\cdot 10^{-3}$\\
						$I_8$~(m$^4$)&Gaussian&$0.0214$&$2.6\cdot 10^{-3}$\\
						$I_9$~(m$^4$)&Gaussian&$0.026$&$3.03\cdot 10^{-3}$\\
						$I_{10}$~(m$^4$)&Gaussian&$0.0108$&$2.6\cdot 10^{-3}$\\
						$I_{11}$~(m$^4$)&Gaussian&$0.0141$&$3.46\cdot 10^{-3}$\\
						$I_{12}$~(m$^4$)&Gaussian&$0.0233$&$5.62\cdot 10^{-3}$\\
						$I_{13}$~(m$^4$)&Gaussian&$0.026$&$6.49\cdot 10^{-3}$\\
						$A_{14}$~(m$^2$)&Gaussian&$0.313$&$0.0558$\\
						$A_{15}$~(m$^2$)&Gaussian&$0.372$&$0.0744$\\
						$A_{16}$~(m$^2$)&Gaussian&$0.506$&$0.093$\\
						$A_{17}$~(m$^2$)&Gaussian&$0.558$&$0.112$\\
						$A_{18}$~(m$^2$)&Gaussian&$0.253$&$0.093$\\
						$A_{19}$~(m$^2$)&Gaussian&$0.291$&$0.102$\\
						$A_{20}$~(m$^2$)&Gaussian&$0.373$&$0.121$\\
						$A_{21}$~(m$^2$)&Gaussian&$0.419$&$0.195$\\
						\hline
					\end{tabular}
				\end{minipage}
		}}		
	\end{tabular}
	\caption{\emph{Five-story frame}: Problem setup with (a) static system, (b) a table of element properties and (c) a table of the parameter marginal distributions. The element properties are listed as eight sets of Young's moduli $E$, second moment of area $I$ and area $A$ for each element type $B_i$ and $C_i$ with $i\in\{1,\cdots,4\}$. The parameter marginals are given per input parameter where $\mu$ and $\sigma$ denote the mean and standard deviation respectively. For the truncated Gaussians those moments apply to their untruncated variants. \label{fig:Ex3:setup}}	
\end{figure}

In this example we consider the five-story structural frame described in Figure~\ref{fig:Ex3:setup}. This example has been investigated before in \citet{LiuPL91, Wei2007, Blatman2010Adaptive}. It consists of eight different structural elements with uncertain dimensions and material properties. The structure is subject to uncertain, horizontal loads $P_1$, $P_2$ and $P_3$. 

The input parameters are collected in a random vector $\ve{X}=\{P_1,\cdots,A_{21}\}$ and sorted according to Figure~\subref{fig:Ex3:setup:parameters}. These parameters are not mutually independent. Instead, we impose a \emph{Gaussian copula} \citep{Nelsen2006} such that the joint input distribution function can be written as
\begin{equation}
	F_{\vX}(\vx) = \Phi_{21}(\Phi^{-1}(F_{X_1}(x_1)),\cdots,\Phi^{-1}(F_{X_{21}}(x_{21}));\ve{R}).
\end{equation} 

In this expression $F_{X_i}$ are the \emph{cumulative distribution functions} (CDF) of the parameters as listed in Figure~\subref{fig:Ex3:setup:parameters}, $\Phi_{21}(\ve{U};\ve{R})$ is the CDF of a $21$-variate Gaussian distribution with mean $\ve{0}$ and correlation matrix $\ve{R}\in\mathbb{R}^{21\times 21}$ and $\Phi^{-1}$ is the inverse CDF of the standard Gaussian distribution. Correlation is assumed only between some parameters, as reflected by the entries of the correlation matrix $\ve{R}$ detailed below
\begin{description}
	\item[Geometrical properties] The geometrical properties are assumed to be correlated with $\rho_{A_i,I_j}=\rho_{I_i,I_j}=\rho_{A_i,A_j}=0.13$. For geometrical properties belonging to the same element type, a stronger correlation of $\rho_{A_i,I_{i+8}}=0.95,i=6\cdots,13$ is assumed.
	\item[Material properties] The two Young's moduli are correlated with $\rho_{E_4,E_5}=0.9$.
\end{description}

The quantity of interest is the top-floor displacement $u(\vX)$. This displacement can be computed as the solution of an elastic structural mechanics problem that is approximately solved with the linear finite element method. We consider the problem of determining the probability that this top-floor displacement exceeds a threshold of $u_{\mathrm{th}}=9$~cm. This yields the limit-state function
\begin{equation}
	\label{ex3:ls}
	g(\vX) =u_{\mathrm{th}} - u(\vX).
\end{equation}

This leads to a reference failure probability of $P_f=1.49\cdot 10^{-6}$ ($\beta=4.67$) computed with Monte Carlo simulation and $N=10^8$ evaluations of the limit-state function. 

We perform $50$ independent runs of SSER on this problem, with different random seeds. The number of refinement samples is set to $\NRefine=40$ and the maximum polynomial degree for the residual expansions to $p_{\mathrm{max}}=4$. The domain-wise accuracy is estimated with $B=500$ bootstrap replications. 

\begin{figure}
	\centering
	\subfloat[Single run]{
		\begin{minipage}{0.49\linewidth}
			\includegraphics[width=\linewidth,clip=true,trim=0 0 0 0]{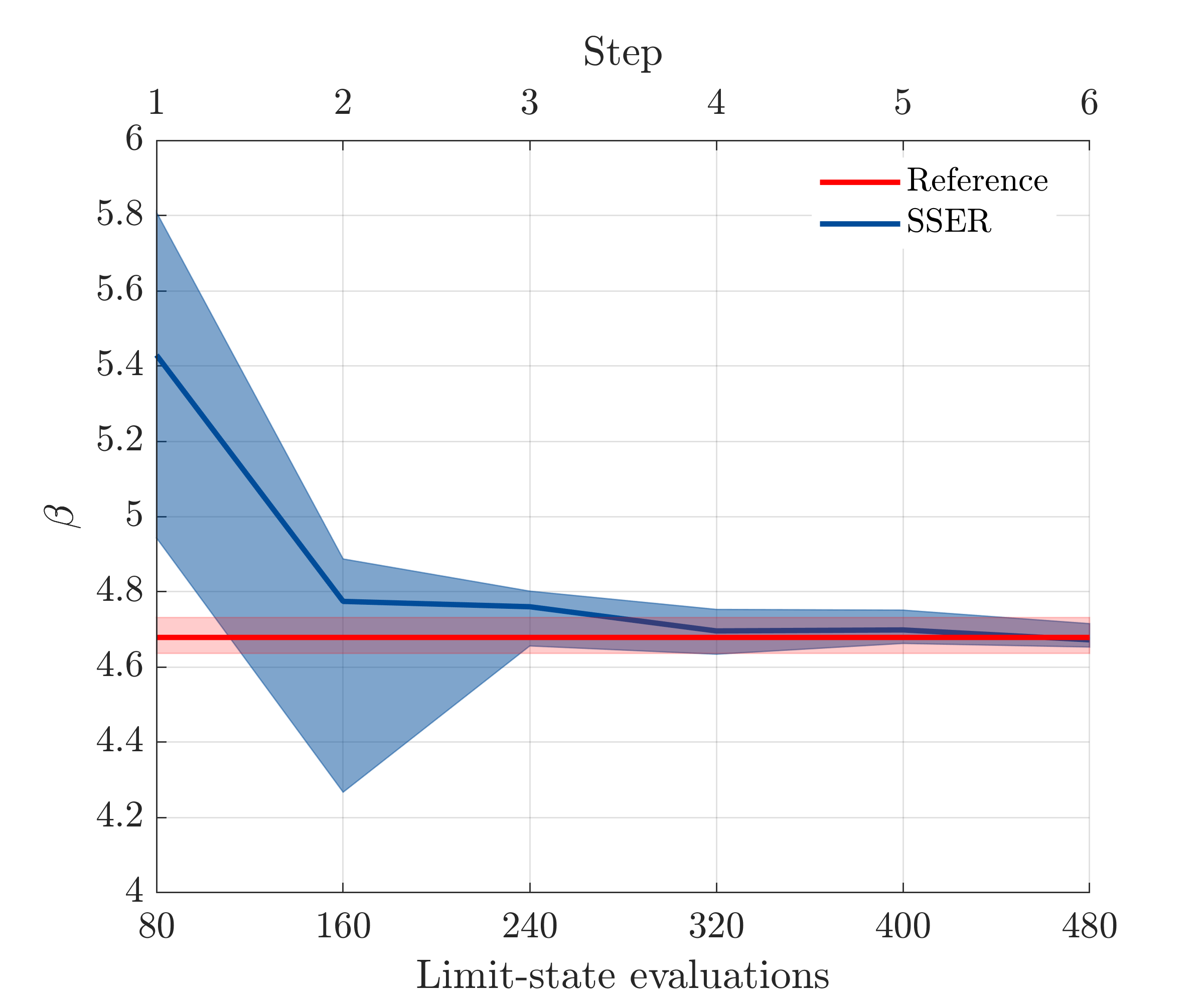}
		\end{minipage}
		\label{fig:Ex3:convergence:single}
	}%
	\subfloat[$50$ independent runs]{
		\begin{minipage}{0.49\linewidth}
			\includegraphics[width=\linewidth,clip=true,trim=0 0 0 0]{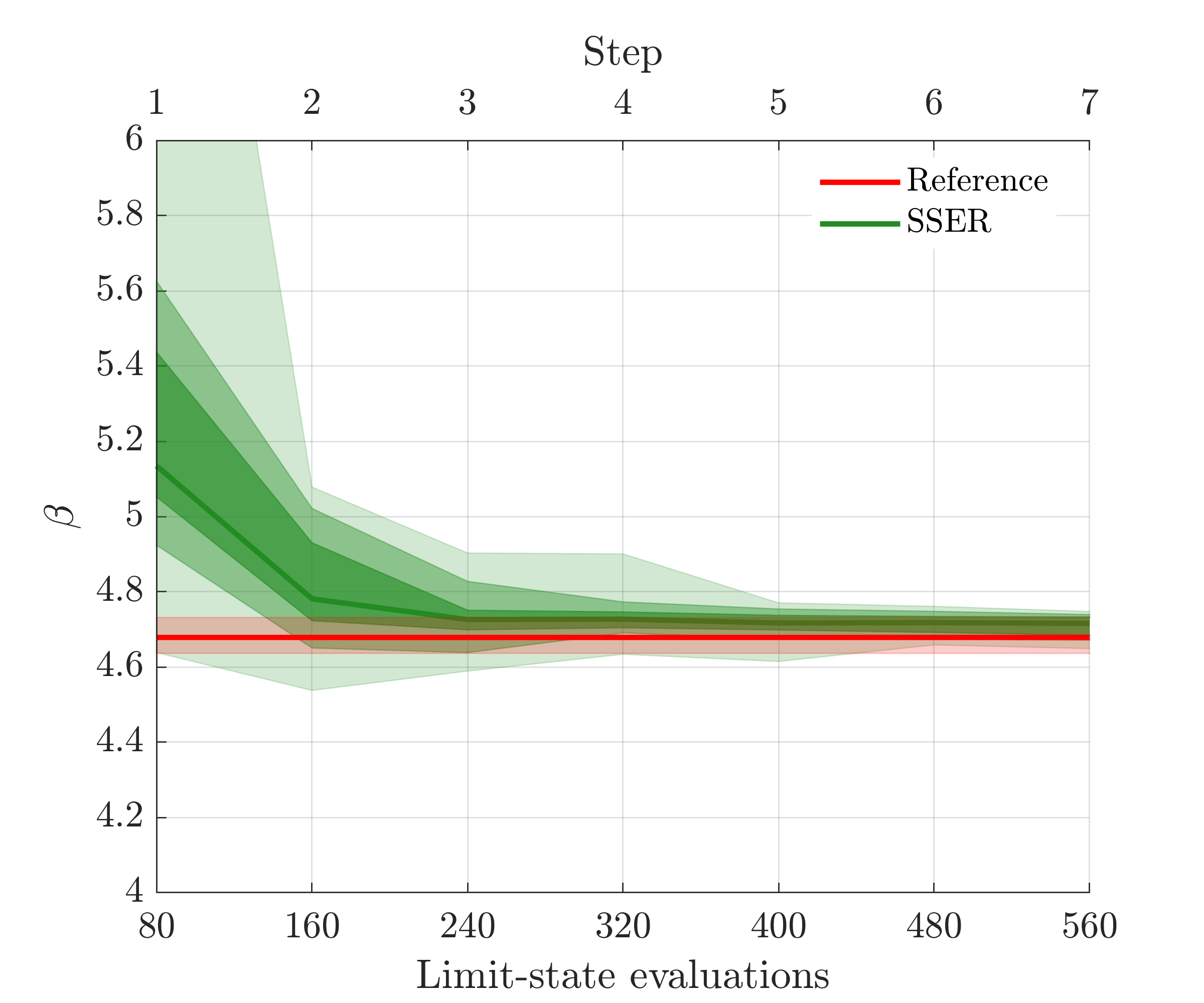}
		\end{minipage}
		\label{fig:Ex3:convergence:replications}
	}
	\caption{\emph{Five-story frame}: Convergence to the reference reliability index $\beta$ as a function of limit-state evaluations and algorithm steps. The $95\%$ confidence intervals in Figure~\protect\subref{fig:Ex3:convergence:single} are based on the bootstrap replications. This figure shows the convergence until the stopping criterion in Eq.~\eqref{eq:stoppingCriterion} is met at Step~$6$ of SSER. The bounds in Figure~\protect\subref{fig:Ex3:convergence:replications} are the $90\%$, $75\%$ and $50\%$ confidence intervals from the mean predictions of the independent SSER runs. \label{fig:Ex3:convergence}}	
\end{figure}

The convergence to the reference reliability index for a single run of SSER and all $50$ runs is shown in Figure~\ref{fig:Ex3:convergence}. The median number of steps required to reach the stopping criterion is $3$, corresponding to $\underbrace{2\NRefine}_{\text{initial}} + (2\cdot 2) \NRefine=240$ limit-state evaluations. The stopping criterion for the shown run in Figure~\subref{fig:Ex3:convergence:single} is met at Step~$6$.

\subsection{Plate with a hole}
\label{sec:Applications:PlateWithHole}

\begin{figure}
	\centering
	\includegraphics[width=0.6\linewidth,clip=true,trim=0 0 0 0]{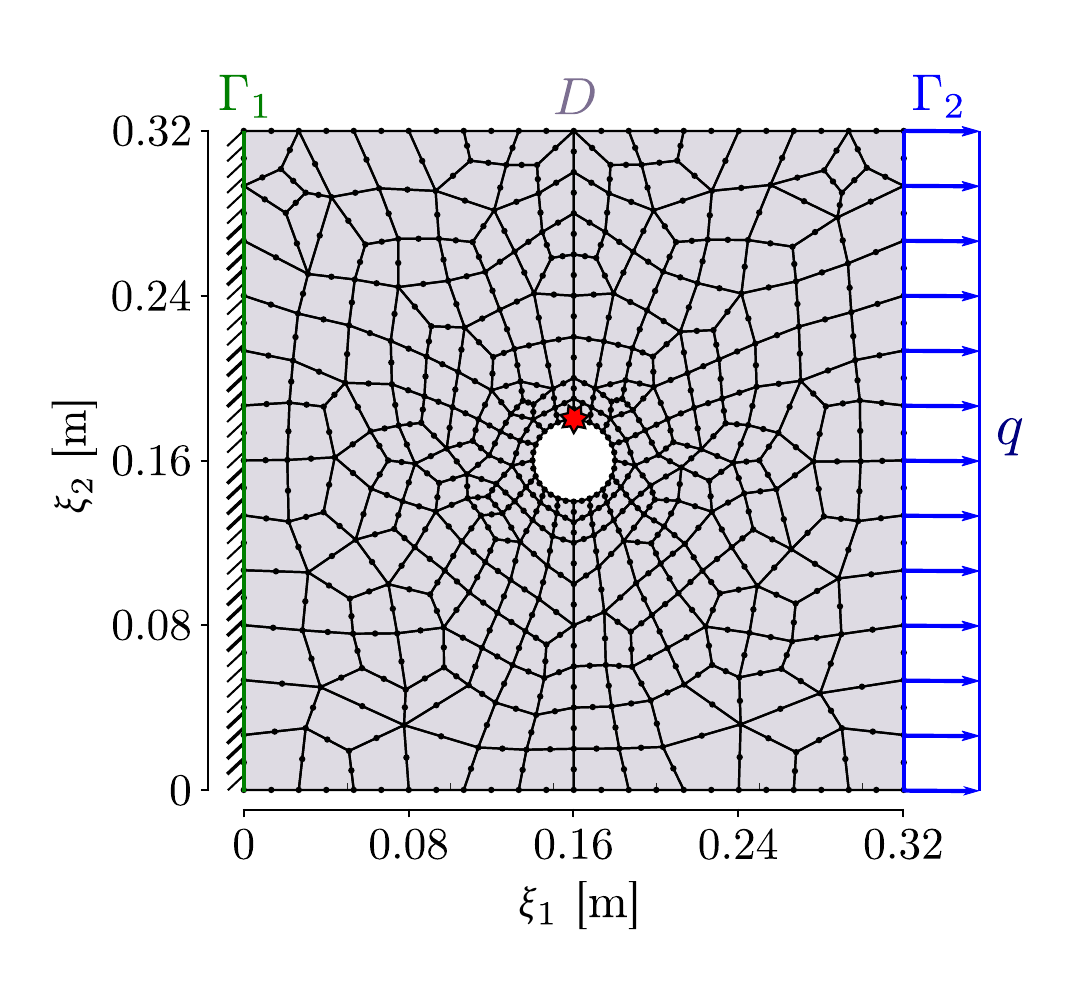}
	\caption{\emph{Plate with a hole}: Finite element discretization with $282$ eight-node serendipity quadrilateral elements \citep{Uribe2020}. The red star \textcolor{red}{$*$} marks the location of the control node for the stress exceedance criterion in Eq.~\eqref{ex4:ls}. \label{fig:Ex4:setup}}	
\end{figure}

This last example was previously analysed in \citet{Uribe2020} and its description here follows closely the description there. Consider a two-dimensional, square steel plate under plane stress conditions with a hole in the middle (see Figure~\ref{fig:Ex4:setup}). The plate is defined on a square domain $\mathcal{D}_{\ve{\xi}}$ with side lengths $0.32$~m, thickness $0.01$~m and a centre hole of radius $0.02$~m. The coordinates in the domain are parametrized by $\ve{\xi}=[\xi_1,\xi_2]\in\mathcal{D}_{\ve{\xi}}$. The plate is assumed to deform elastically with an isotropic Young's modulus $E(\ve{\xi})$. Neglecting body-forces results in the following system of partial differential equations (PDEs)
\begin{equation}
	\label{eq:Ex4:PDEs}
	\frac{E(\ve{\xi})}{2(1+\nu)}\nabla^2 \ve{u}(\ve{\xi}) + \frac{E(\ve{\xi})}{2(1-\nu)}\nabla(\nabla\ve{u}(\ve{\xi}))^{\intercal} = 0,
\end{equation}
where $\ve{u}(\ve{\xi})$ is the 2D displacement field and $\nu=0.29$ is the steel Poisson's ratio. As sketched in Figure~\ref{fig:Ex4:setup} a Dirichlet boundary condition is applied on the left domain border such that $\ve{u}(\ve{\xi})=0$ for $\ve{\xi}\in\Gamma_1$. Additionally, a Neumann boundary condition is applied on the right domain border such that $\sigma_{11}(\ve{\xi})=q$ for $\ve{\xi}\in\Gamma_2$, where $\sigma_{11}$ is the stress component in the first direction.

The Young's modulus $E(\ve{\xi})$ is modelled as lognormal random field with mean $\mu_E=2\cdot 10^{5}$~MPa and standard deviation $\sigma_{E}=3\cdot 10^4$~MPa. We use an isotropic exponential kernel $k(\ve{\xi},\ve{\xi}')=\exp{(-\ell^{-1}\vert\vert\ve{\xi}-\ve{\xi}'\vert\vert_2)}$ as the autocorrelation function of the underlying Gaussian random field with correlation length $\ell=0.04$~m. To parametrize this random field, we resort to the Karhunen-Lo{\`e}ve expansion \cite{Ghanem} and write
\begin{equation}
	E(\ve{\xi})\approx\widehat{E}(\ve{\xi},\ve{\theta})\eqdef\exp{\left(\mu_{E'}+\sum_{k=1}^K\sqrt{\lambda_k}\varphi_k(\ve{\xi})\theta_k\right)},
\end{equation}
where $\lambda_k$ and $\varphi_k(\ve{\xi})$ are the ordered (\ie $\lambda_k\ge\lambda_{k+1}$) eigenvalues and eigenfunctions of the covariance operator $C(\ve{\xi},\ve{\xi}')\eqdef\sigma^2_{E'}k(\ve{\xi},\ve{\xi}')$. The mean and variance of the underlying Gaussian field are $\mu_{E'}$ and $\sigma^2_{E'}$ respectively and $\ve{\theta}\in\mathbb{R}^k$ denotes the standard Gaussian vector of random coefficients. The expansion is truncated at $K=868$ terms, which account for $92.5\%$ of the spatial average of the variance of the Gaussian random field. The eigenpairs are estimated with the Nyström method based on $100$ Gauss-Legendre points in each direction, disregarding points inside the hole.

Additionally, we model the applied external load as a normally distributed random variable with mean $\mu_q=60$~MPa and standard deviation $\sigma_q=12$~MPa such that $q\sim\mathcal{N}(\mu_q,\sigma_q)$. Assuming further independence between the standard normal vector $\ve{\theta}$ and $q$, the probabilistic input vector of this problem is gathered in $\vX = (\ve{\theta}, q)$. 

The quantity of interest is the \emph{principal stress} at a \emph{control point} marked by a red star in Figure~\ref{fig:Ex4:setup} defined as
\begin{equation}
	\sigma_{\color{red}*}\eqdef (\sigma_{11}+\sigma_{22})/2 + \sqrt{((\sigma_{11}-\sigma_{22})/2)^2+\tau^2_{12}}.
\end{equation}

It depends on the stress field $\ve{\sigma}(\ve{\xi})$ at the control point that is obtained from the elastic constitutive equations after solving Eq.~\eqref{eq:Ex4:PDEs} with the finite element method for the displacement field $\ve{u}(\ve{\xi})$. Failure in this problem is defined as this principle stress exceeding $\sigma_{\mathrm{th}}=320$~MPa, which can be formalized in the following limit-state function:
\begin{equation}
	\label{ex4:ls}
	g(\vX) = \sigma_{\mathrm{th}} - \sigma_{\color{red}*}(\vX).
\end{equation}

Due to the complexity of this problem, it is not possible to obtain a reference solution with Monte Carlo simulation in a reasonable time. We use instead directly the reference solution from \citet{Uribe2020} that was computed with $100$ independent subset simulation \citep{Au2001} runs with $N=3{,}000$ samples per level and $6$ levels, leading to a reference failure probability of $P_f=3.75\cdot 10^{-6}$ ($\beta=4.48$).

On this problem, we perform $10$ independent runs of SSER with different random seeds. The number of refinement samples was set to $\NRefine=100$ and the maximum polynomial degree for the residual expansions to $p_{\mathrm{max}}=4$. To allow PCEs to be constructed efficiently in such high dimensions, it was necessary to restrict the maximum interactions of polynomial basis functions to $1$ \citep{UQdoc_14_104}. The domain-wise accuracy was estimated with $B=100$ bootstrap replications. 

\begin{figure}
	\centering
	\subfloat[Single run]{
		\begin{minipage}{0.49\linewidth}
			\includegraphics[width=\linewidth,clip=true,trim=0 0 0 0]{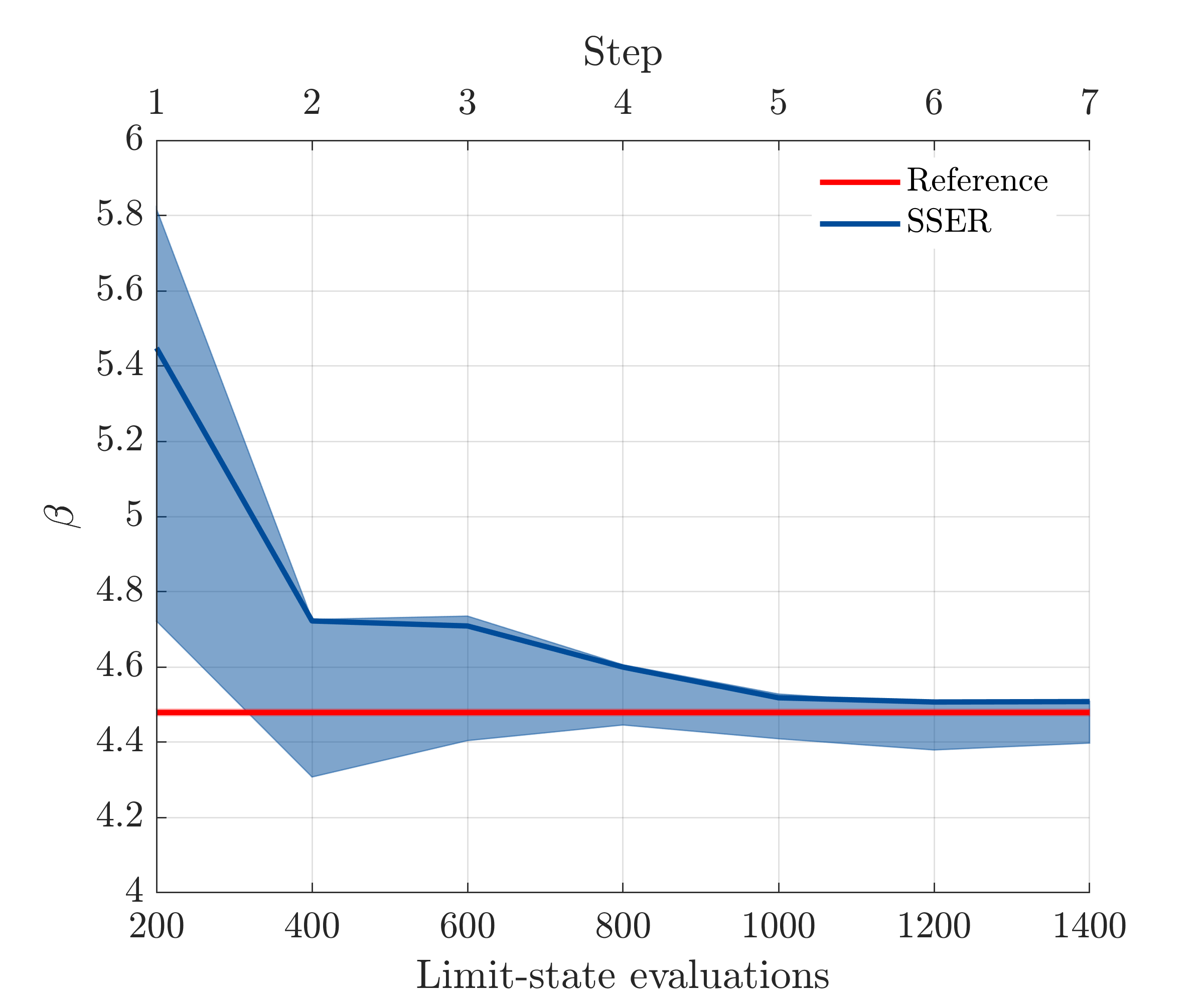}
		\end{minipage}
		\label{fig:Ex4:convergence:single}
	}%
	\subfloat[$10$ independent runs]{
		\begin{minipage}{0.49\linewidth}
			\includegraphics[width=\linewidth,clip=true,trim=0 0 0 0]{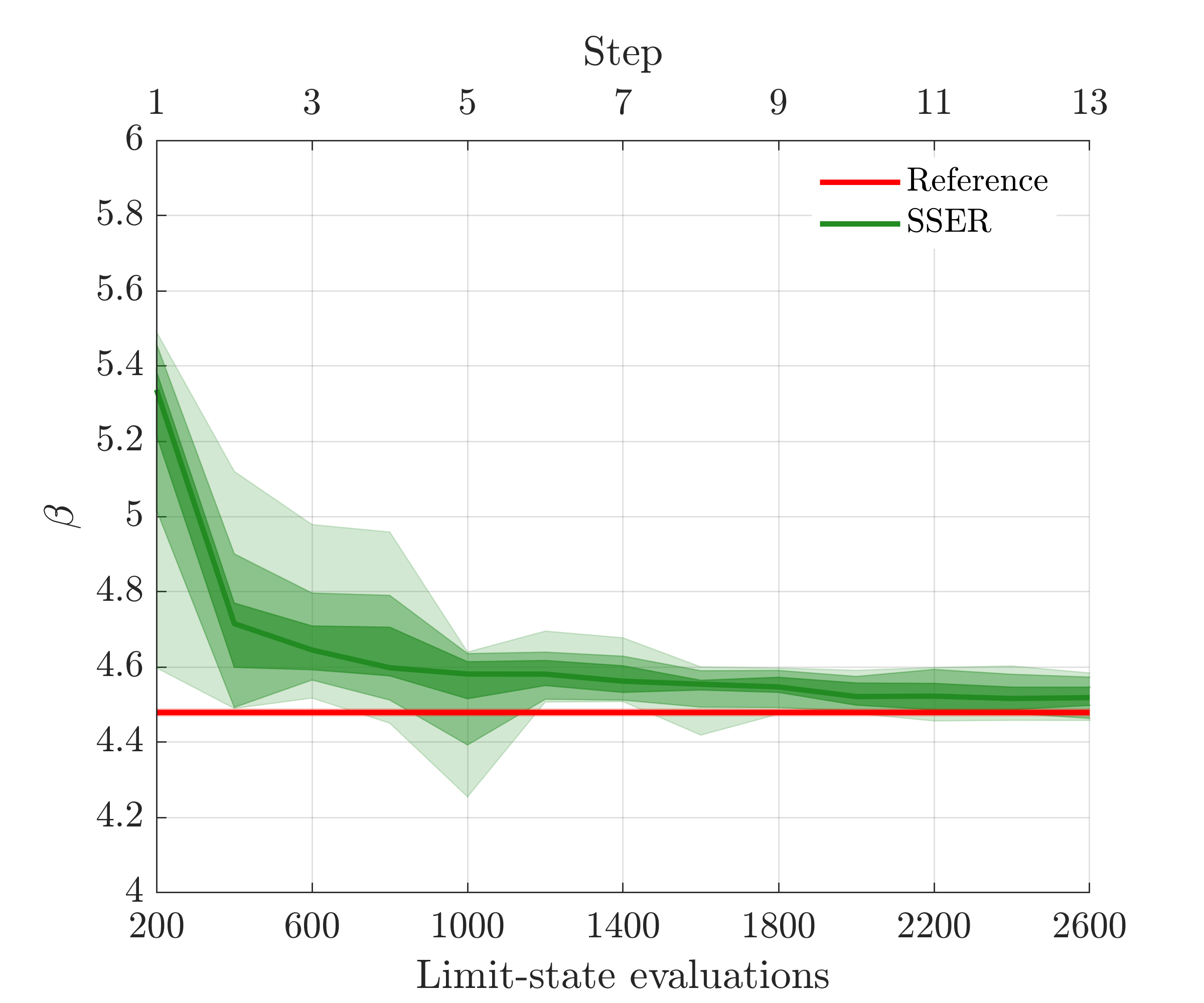}
		\end{minipage}
		\label{fig:Ex4:convergence:replications}
	}
	\caption{\emph{Plate with a hole}: Convergence to the reference reliability index $\beta$ as a function of limit-state evaluations and algorithm steps. The $95\%$ confidence intervals in Figure~\protect\subref{fig:Ex4:convergence:single} are based on the bootstrap replications. This figure shows the convergence until the stopping criterion in Eq.~\eqref{eq:stoppingCriterion} is met at Step~$7$ of SSER. The bounds in Figure~\protect\subref{fig:Ex4:convergence:replications} are the $90\%$, $75\%$ and $50\%$ confidence intervals from the mean predictions of the independent SSER runs. \label{fig:Ex4:convergence}}	
\end{figure}

The convergence of SSER to the reference reliability index is shown in Figure~\ref{fig:Ex4:convergence}. The median number of steps required to reach the stopping criterion is $6$, corresponding to $\underbrace{2\NRefine}_{\text{initial}} + (5\cdot 2) \NRefine=1{,}200$ limit-state evaluations.

\section{Conclusions}
\label{sec:Conclusions}
Leveraging the flexibility of the recently proposed stochastic spectral embedding formalism \citep{Marelli2020}, we show that the adaptive sequential partitioning approach introduced in \citep{Wagner2020JCP} can be efficiently modified to an active learning reliability method. The introduced modifications pertain to the refinement domain selection, partitioning and sample enrichment strategies. 

The \rev{SSER} algorithm is shown to accurately estimate failure probabilities spanning multiple orders of magnitude at a fraction of the cost of plain Monte Carlo simulation. The method is suitable for identifying multiple failure regions (\emph{four-branch function}, Section~\ref{sec:Applications:FourBranch}) and is able to deal with problems that involve non-smooth limit-state functions (\emph{piecewise linear function}, Section~\ref{sec:Applications:PiecewiseLinear}). It also performs reasonably well in moderate-dimensions with dependent input vectors (\emph{five-story frame}, Section~\ref{sec:Applications:5StoryFrame} and is competitive to a state-of-the-art reliability method iCEred \citep{Uribe2020} on a high-dimensional example (\emph{plate with a hole}, Section~\ref{sec:Applications:PlateWithHole}).

\rev{In general, the efficiency of SSER depends on two factors: (1) whether the algorithm identifies all failure modes and (2) whether the chosen spectral expansion technique can efficiently approximate $g$ at the terminal domains that lie in proximity of the limit-state surface. Regarding the first factor, the refinement domain selection criterion in Eq.~\eqref{eq:algo:refinement} should guarantee that SSER eventually identifies all modes of failure (\eg \emph{four-branch function} and \emph{piecewise linear function}). However, in problems with a large number of disjoint limit-states, this sequential process might require a considerable number iterations and consequently model evaluations. Also, if the stopping criterion is triggered before all limit-states have been identified, the total failure probability might be underestimated. The second factor depends on $g$'s complexity at the terminal domains in proximity of the limit-state surface. Generally, spectral expansion techniques are known to converge slowly in highly non-linear and high-dimensional problems. Hence, we expect a decrease in efficiency of the algorithm in such problems. As an example, SSER is expected to perform poorly in first excursion problems that are both high-dimensional and highly non-linear in the limit state-function in proximity of the limit-state surface.}

In some examples (most notably in Section~\ref{sec:Applications:PlateWithHole}) there persists a small bias towards a higher reliability index compared to the reference solution. This means that the method systematically misses small parts of failure regions. A possible explanation for this is the partitioning strategy that `unknowingly' cuts away undiscovered parts of failure regions and incorrectly classifies them as safe. This process is facilitated by the boundary discontinuities and will be explored in future works. 

An interesting side result of the proposed method is the domain partition that hints at the location of failure domains. This information, in conjunction with the available \rev{conditional} failure probabilities, could be directly used for design purposes. 

The modularity of the sequential partitioning algorithm for constructing SSEs makes it well suited to for future modifications. Possible improvements could focus on different spectral expansion or even surrogate modelling techniques, as well as simultaneous partitioning in multiple dimensions.
%
%
A particularly interesting idea is the introduction of rotated domain partitions by means of partial least squares \cite{Papaioannou2019}. These partitions could be more flexibly tuned to fit the limit-state surface and thereby increase the efficiency of SSER.

\section*{Acknowledgements}
The PhD thesis of the first author is supported by ETH grant \#44 17-1.


\bibliographystyle{chicago}
\bibliography{References}
\end{document}